\documentclass{article} 
\usepackage{iclr2025_conference,times}

\usepackage{amsmath,amsfonts,bm}

\def\eqref#1{equation~\ref{#1}}

\def\1{\bm{1}}

\DeclareMathAlphabet{\mathsfit}{\encodingdefault}{\sfdefault}{m}{sl}
\SetMathAlphabet{\mathsfit}{bold}{\encodingdefault}{\sfdefault}{bx}{n}

\usepackage{hyperref}
\usepackage{url}

\usepackage{colortbl}
\definecolor{maroon}{cmyk}{0,0.87,0.68,0.32}
\usepackage{tikz}
\usepackage{filecontents}
\usepackage{pgfplots}
\usepackage{pgfplotstable}
\usetikzlibrary{positioning, shapes.geometric, shapes.arrows, decorations.pathreplacing, shapes.misc,calc,fit,backgrounds,svg.path}

\pgfplotsset{
    discard if not/.style 2 args={
        x filter/.code={
            \edef\tempa{\thisrow{#1}}
            \edef\tempb{#2}
            \ifx\tempa\tempb
            \else
                
            \fi
        }
    }
}

\makeatletter
\pgfplotstableset{
    discard if not/.style 2 args={
        row predicate/.code={
            \def\pgfplotstable@loc@TMPd{\pgfplotstablegetelem{##1}{#1}\of}
            \expandafter\pgfplotstable@loc@TMPd\pgfplotstablename
            \edef\tempa{\pgfplotsretval}
            \edef\tempb{#2}
            \ifx\tempa\tempb
            \else
                \pgfplotstableuserowfalse
            \fi
        }
    }
}
\makeatother

\usepackage{xparse}
\NewDocumentCommand\Cycle{O{} m m m O{} m}{%
  \draw[#1](#2.{#3+asin(#6/(#4*1.41))}) arc (180+#3-45:180+#3-45-270:#6/2) #5;
}

\usepackage{hyperref}

\usepackage{amsmath,amssymb,amsfonts}
\usepackage{algorithmic}
\usepackage{algorithm}
\usepackage{graphicx}
\usepackage{textcomp}
\usepackage{xcolor}
\usepackage{multirow}
\usepackage{xspace}
\usepackage{subfig}
\usepackage{standalone}
\usepackage{comment}

\usepackage[noabbrev,capitalise]{cleveref}

\usepackage{pgfplots, pgfplotstable}
\usepackage{float}
\usepackage{comment}

\usepackage[boldmath]{numprint} 
\npdecimalsign{.}
\nprounddigits{3}

\usepackage{array}
\newcolumntype{H}{>{\setbox0=\hbox\bgroup}c<{\egroup}@{}}  

\usepackage{siunitx}

\usepackage{subcaption}

\usepackage{wrapfig}

\newcommand{\contentwidth}{\columnwidth}

\newcommand{\modelfull}{Time-aware Diffusion Paint\xspace}
\newcommand{\model}{TD-Paint\xspace}
\newcommand{\modelrepaint}{RePaint\xspace}
\newcommand{\modelvanilla}{RePaint-1\xspace}
\newcommand{\modelmcg}{MCG\xspace}
\newcommand{\modelmat}{MAT\xspace}
\newcommand{\modellama}{LaMa\xspace}
\newcommand{\modelldm}{LDM\xspace}
\newcommand{\modelcopaint}{CoPaint\xspace}
\newcommand{\modelbld}{BLD\xspace} 
\newcommand{\modelunipaint}{Uni-paint\xspace} 
\newcommand{\modelpowerpaint}{PowerPaint\xspace}
\newcommand{\modelcontrolnet}{ControlNet\xspace}

\newcommand{\celebahq}{CelebA-HQ\xspace} 
\newcommand{\imagenet}{ImageNet1K\xspace} 
\newcommand{\places}{Places2\xspace}

\newcommand{\tmap}{\ensuremath{\tau}\xspace}

\newcommand{\know}{\text{$\oplus$}\xspace}
\newcommand{\unknow}{\text{$\ominus$}\xspace}

\newcommand{\nbsteps}{NBS\xspace}

\newcommand{\ssim}{SSIM\xspace}
\newcommand{\lpips}{LPIPS\xspace}
\newcommand{\kid}{KID\xspace}

\newcommand{\wide}{\textit{Wide}\xspace}
\newcommand{\narrow}{\textit{Narrow}\xspace}
\newcommand{\twotime}{\textit{Super-Resolve 2x}\xspace}
\newcommand{\altline}{\textit{Altern. Lines}\xspace}
\newcommand{\half}{\textit{Half}\xspace}
\newcommand{\expand}{\textit{Expand}\xspace}

\newcommand{\mto}{\text{Ours}\xspace}
\newcommand{\mtl}{\text{LAMA}\xspace}
\newcommand{\mtol}{\text{Ours+LAMA}\xspace}

\definecolor{repaintfrd}{HTML}{2f9e44}
\definecolor{repaintbck}{HTML}{da77f2}

\newcounter{cpt}
\usepackage{multirow}
\usepackage{tikz}

\title{\model: Faster Diffusion Inpainting Through Time-aware Pixel Conditioning}

\author{Tsiry Mayet \\
INSA Rouen Normandie, LITIS UR 4108, \\
F-76000 Rouen, France \\
\And
Pourya Shamsolmoali \\
University of York, United Kingdom \\
East China Normal University, China \\
\And
Simon Bernard \\
Université Rouen Normandie, LITIS UR 4108, \\
F-76000 Rouen, France \\
\And
Eric Granger \\
LIVIA, Dept. of Systems Engineering, \\
ETS Montreal, Canada \\
\And
Romain Hérault \\
Université Caen Normandie, CNRS, GREYC UMR6072, \\
F-14000, Caen, France \\
\And
Clément Chatelain \\
INSA Rouen Normandie, LITIS UR 4108, \\
F-76000 Rouen, France \\
}

\iclrfinalcopy 
\begin{document}

\maketitle

\begin{abstract}
Diffusion models have emerged as highly effective techniques for inpainting, however, they remain constrained by slow sampling rates. While recent advances have enhanced generation quality, they have also increased sampling time, thereby limiting scalability in real-world applications. We investigate the generative sampling process of diffusion-based inpainting models and observe that these models make minimal use of the input condition during the initial sampling steps. As a result, the sampling trajectory deviates from the data manifold, requiring complex synchronization mechanisms to realign the generation process. To address this, we propose \modelfull(\model), a novel approach that adapts the diffusion process by modeling variable noise levels at the pixel level. This technique allows the model to efficiently use known pixel values from the start, guiding the generation process toward the target manifold. By embedding this information early in the diffusion process, \model significantly accelerates sampling without compromising image quality. Unlike conventional diffusion-based inpainting models, which require a dedicated architecture or an expensive generation loop, \model achieves faster sampling times without architectural modifications. Experimental results across three datasets show that \model outperforms state-of-the-art diffusion models while maintaining lower complexity.
Github code: \href{https://github.com/MaugrimEP/td-paint}{https://github.com/MaugrimEP/td-paint}
\end{abstract}

\section{Introduction}
\begin{figure}%
    \centering%
        \includegraphics[width=0.49\textwidth]{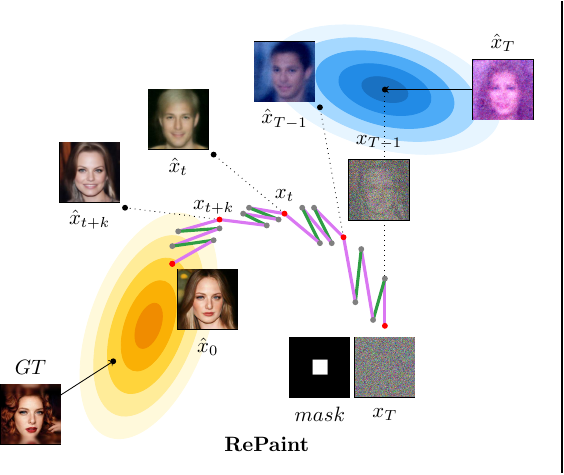}%
        \includegraphics[width=0.3\textwidth]{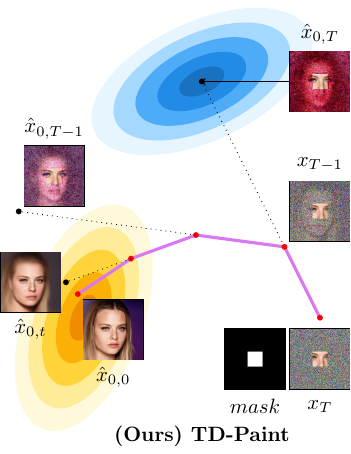}%
        \caption{%
        \textbf{Comparison of noisy-condition models (e.g., \modelrepaint) and \model generation processes.}
        Reverse denoising steps ($p_\theta(x_{t-1} | x_{t})$) are depicted \textcolor{repaintbck}{in purple lines ---}, forward noising steps ($ q(x_t |x_{t-1})$) are shown \textcolor{repaintfrd}{in green lines ---}.
        Here, $x_t$ represents the input to the diffusion process at step $t$, $mask$ denotes the conditional area (shown in white) and the region to be generated (shown in black), and $\hat x_{t}$ represents the model's prediction at diffusion step $t$.
        \textbf{(left) \modelrepaint\citep{lugmayr2022repaint} generation process}. \modelrepaint applies a cycle of reverse and forward diffusion steps. It can be observed that the intermediate steps of the generation process lack consistency, changing from a man with dark hair at $\hat x_{T-1}$ to a man with blond hair at $\hat x_{t}$ to a woman at $\hat x_{t+k}$. These changes occur due to the synchronization process, where the initially predicted images $\hat x_{t}$ are not well aligned with the given condition.%
        \textbf{(right) \model generation process}. In comparison, \model can use a clean condition from the beginning of the inpainting process, resulting in a faster and more stable process.
        Note how the intermediate \model steps are consistent from one to another.%
        }%
\label{fig:repaintvsours}%
\end{figure}%
Given an image and a binary mask, image inpainting aims to generate the missing region while preserving the semantics of the visible region. This task is challenging because the generated content must not only be coherent with the existing parts of the image but also appear realistic. Additionally, the generation process should be stochastic to produce diverse outputs. An effective inpainting model must also address variations in mask shape and size. Generalizing to unseen masks during training and accurately filling large missing regions further complicates the task.

Diffusion models have shown significant success as generative models \citep{dhariwal2021diffusion, rombach2022highresolution}, by approximating the distribution of real images through a fixed Markov chain that transforms Gaussian noise into the real image distribution. During training, a forward diffusion process gradually adds noise to an image, and the model is trained to reverse this process, learning to denoise and recover the original image distribution.
During generation, the backward diffusion iteratively removes noise from an initial Gaussian noise image. The trained model predicts and removes noise at each step, gradually refining the image until a photorealistic result is achieved.

Methods proposed in the literature have investigated using the standard diffusion model \citep{lugmayr2022repaint,chung2022improving} for inpainting by combining the noisy condition with the current generation at each step.
This technique has its limitations. While the textures match, it creates disharmony between the conditioning and generation parts.
This disharmony comes from the fact that, during the early steps of the generation process, the condition contains a lot of noise that the model cannot leverage. Therefore, the generation moves away from the intended semantics and produces unsatisfactory results.
An illustration of such inpainting can be found in \cref{fig:repaintvsours}(left).

To address this limitation, \modelrepaint~\citep{lugmayr2022repaint} introduced a resampling mechanism that repeats the diffusion steps multiple times, enabling the model to synchronize condition and generation better. Although \modelrepaint produces highly faithful images, this resampling significantly slows the generation process. Indeed, \modelrepaint requires approximately 5k steps to generate a single image, increasing time complexity.  An illustration of \modelrepaint's generative process is shown in \cref{fig:repaintvsours} (left).
Other approaches introduce additional constraints at each diffusion step\citep{chung2022improving,li2023image}. For example, \citep{chung2022improving} integrates a correction mechanism that encourages the diffusion path to remain close to the data manifold. This is achieved by minimizing the reconstruction error of the known image region relative to the unknown region.

In contrast to models that use a noisy condition\citep{lugmayr2022repaint,chung2022improving}, our Time-aware Diffusion Paint (\model) approach integrates the currently generated sample with a clean condition. Our method uses semantic information from the beginning of the generation process, resulting in a more efficient and cost-effective approach.
To achieve this, \model uses time conditioning derived from the standard formulation of diffusion models. Instead of using a single scalar $t$ for the entire image, \model assigns a unique $t$ value to each pixel. The known image region is assigned a smaller $t$, indicating lower noise levels. In contrast, the region to be generated is assigned a $t$ value proportional to the current step in the generation process. An illustration of \model's generative process is provided in \cref{fig:repaintvsours}(right).

\noindent\textbf{Our main contributions are summarized as follows:}\\
    \noindent\textbullet \  We propose a novel noise modeling paradigm for diffusion models that allows for integrating varying noise levels into the input of diffusion models for the known and unknown regions. \model exploits time conditioning in diffusion models to achieve faster and higher-quality generation.\\ 
    \noindent\textbullet \ An extensive set of experiments on the challenging \celebahq, \imagenet and Places2 datasets demonstrates that our \model, not only outperforms state-of-the-art diffusion-based models but also surpasses other inpainting methods, including those based on CNNs and transformers. Additionally, the results indicate \model is a more cost-effective solution for diffusion models.
\section{Related Works}\label{related_works}

Inpainting aims to fill in a missing part of the image realistically. Traditional inpainting methods try to combine techniques to propagate texture and structural information onto the missing parts \citep{1323101}. Some algorithms \citep{10.1007/978-3-540-24671-8_17} use large image datasets and assume that the possible semantic space for missing regions is limited.
In recent years, deep learning models for inpainting have made impressive progress using two types of generative models, VAEs \citep{kingma2022autoencoding} and GANs \citep{goodfellow2014generative, mirza2014conditional}.

\noindent\textbf{(a) Single-Stage Inpainting:} 
Most single-stage methods use the context encoding setting introduced by Pathak et al. \citep{pathak2016context}, with an encoder-decoder setup. A reconstruction loss (L2) ensures global structure consistency, while an adversarial loss ensures the reconstruction is realistic.
Global consistency is an important consideration. CNNs are limited by a receptive field that grows slowly with network layers. Many layers are needed for information to travel from one side of the image to the other.
Dilated convolution \citep{yu2015multi} has been used by \citep{iizuka2017globally} to increase the receptive field. 
Partial convolution \citep{liu2018image} uses mask information to attend only the visible regions.
The pyramid context encoder \citep{zeng2019learning} learns an affinity map between regions in a pyramidal fashion.
Fourier convolution \citep{suvorov2021resolutionrobust} aims to provide a global receptive field to both the inpainting network and the loss function. Fast Fourier Convolutions have an image-wide receptive field, which helps with large missing areas.
Mask-Aware Transformer (MAT) \citep{li2022mat} is a transformer-based architecture that allows the processing of high-resolution images. A customized transformer block considers only valid tokens, and a style manipulation module updates convolution weights with noise to produce diverse outputs.

\noindent\textbf{(b) Progressive Image Inpainting:}
These methods seek to address global consistency by using coarse-to-fine multi-stage approaches.
Multiple generations are possible for a large missing region in single-stage training. Some may have a large pixel-to-pixel distance from the original ground truth, which can be misleading when training models with pixel-wise distance losses.
To address this issue, Yun et al. \citep{yu2018generative} proposed a two-stage generative approach. The first stage produces a coarse output optimized with L1 loss incorporating spatial discounting, while the second stage refines the output further using both global and local critics.
Gated convolution \citep{dauphin2017language} has been used in a coarse-to-fine network to learn valid pixels \citep{yu2019free}.

\noindent\textbf{(c) Prior Knowledge Inpainting:}
These methods leverage and mine information from generative models.
Prior Guided GAN (PGG) \citep{suvorov2021resolutionrobust} uses the latent space of a pre-trained GAN and learns to map masked images to this space using an encoder. A masked image can be mapped to a latent code during inference, and the generator can produce a corresponding inpainted image.
Deep Generative Prior \citep{pan2020exploiting} relaxes the frozen generator assumption of GAN inversion methods and proposes progressively refining each layer.
PSP \citep{richardson2021encoding} uses StyleGAN \citep{karras2019stylebased} latent space to encode an image into its latent space. Inpainting is formulated as a domain translation task performed in the latent space of StyleGAN, removing adversarial components from the training process.

\noindent\textbf{(d) Diffusion Model Inpainting:}
While GANs have recently shown impressive results, most applications are limited to generating a specific domain. In contrast, Diffusion models have gained traction for image generation; Denoising Diffusion Probabilistic Model (DDPM) \citep{ho2020denoising} and Denoising Diffusion Implicit Models (DDIM) \citep{song2022denoising} can generate very diverse and high-quality images.
Some unconditional diffusion models have shown the ability to perform zero-shot inpainting \citep{sohldickstein2015deep,song2021scorebased} but provide only qualitative results.
%
The Pixel Spread Model (PSM) \citep{li2023image} uses a decoupled probabilistic model that combines the efficiency of GAN optimization with the prediction traceability of diffusion models.
Latent Diffusion Models (LDM) \citep{rombach2022highresolution} decouple two tasks: image processing and compression, and the denoising process is learned in latent space instead of pixel space.
Inpainting is performed by encoding the masked input image, downsampling the inpainting mask, and concatenating them as additional conditions to the denoising model.
Designing an image and mask conditional diffusion model requires a special architecture to accept additional input for inpainting, as done in \citep{rombach2022highresolution,li2023image}, and often treats inpainting as a domain translation task.
In contrast, \model does not require any architecture modification by directly combining the masked input and current generation with different noise levels.
Differential Diffusion \citep{levin2023differential} introduces a new approach to soft-inpainting, where both the generated region and the conditional input are modified to ensure coherence across the entire image. This method uses LDM along with a strength map to focus on different image regions during each diffusion step. However, Differential Diffusion operates on noisy conditional inputs.
The Manifold Constrained Gradient (MCG) \citep{chung2022improving} adds a correction term to ensure each sample step remains close to the data manifold, allowing for more stable inpainting.
\modelrepaint \citep{lugmayr2022repaint} which aligns with our approach, combining the noisy conditional region with the current generation, where the diffusion model iteratively updates missing pixels using the surrounding context.
We observe that during the early inpainting step of \modelrepaint, the condition is dominated by noise and does not contain any semantic information.
This causes the model prediction to deviate from the target manifold (see \cref{fig:repaintvsours}(left)).
A resampling mechanism is needed to synchronize the condition and generation regions, allowing semantically corrected images to be produced at the cost of significantly increasing computation time.

Instead of degrading the condition to the same level as the generation, we propose keeping it clean and conditioning the missing part on the known pixels from the beginning of the generation process (see \cref{fig:repaintvsours}(right)).
Although it requires fine-tuning the model, our approach provides much faster sampling by avoiding resampling \citep{lugmayr2022repaint} or additional constraints\citep{chung2022improving}. 
We use temporal information to model the detail in the image.
The condition, containing clean data, is assigned a low noise level while the generation region starts with maximum noise that gradually decreases during the process. This approach allows for a direct combination of the clean and generated regions in the same space without changing the model architecture.
\section{Background On Inpainting Diffusion Models}

\noindent\textbf{Diffusion Models:}
Diffusion models learn a data distribution from a training dataset by inverting a noise process.
During training, the forward diffusion process transforms a data point $x_0$ into Gaussian noise $x_T \sim \mathcal{N}(0, \mathbf{I})$ in $T$ steps by creating a series of latent variables $x_1,...,x_T$ using:
\begin{equation}
    q(x_t | x_{t-1}) = \mathcal{N}(x_t; \sqrt{1- \beta_t} x_{t-1}, \beta_t \mathbf{I}),
\end{equation}
where $\beta_t$ represents the predefined variance schedule. Given $\alpha_t = 1 - \beta_t$, $\bar{\alpha}_t = \prod_{i=1}^t \alpha_i$, and $\epsilon \sim \mathcal{N}(0,\mathbf{I})$, $x_t$ at step $t$ can be marginalized from $x_0$ using the reparametrization trick as follows:
\begin{equation}
    x_t = \sqrt{\bar\alpha_t}x_0 + \sqrt{1-\bar\alpha_t}\epsilon.
\end{equation}
The reverse denoising process $p_\theta(x_{t-1} | x_t, t)$ allows generating from the data distribution by first sampling from $x_T \sim \mathcal{N}(0, \mathbf{I})$ and iteratively reducing the noise in the sequence $x_T,...,x_0$.
The model $\epsilon_\theta(x_t,t)$ is trained to predict the added noise $\epsilon$ to produce the sample $x_t$ at time step $t$. The model is trained using mean square error (MSE):
\begin{equation}
    \mathcal{L} 
    = 
    \mathbb{E}_{
        \epsilon\sim\mathcal{N}(0, \mathbf{I}),
        x_0,
        t
    } 
    \|
        \epsilon_\theta(\sqrt{\bar\alpha_t}x_0 + \sqrt{1-\bar\alpha_t}\epsilon, t) - \epsilon
    \|^2_2.
\end{equation}

\noindent\textbf{RePaint Methodology:}
Given an image $x$ and a binary mask $m$, the goal of inpainting is to generate the missing region $x^{\unknow}$ specified by $x \odot (1-m)$ conditioned on the known region $x^{\know}$ specified by $x \odot m$.
For this, \modelrepaint combines a noisy version of the condition $x_0 \odot m$ with the previous output of the generation process: 
\begin{equation}
    x^{\know}_{t-1} \sim \mathcal{N}\left( \sqrt{\bar\alpha_t}x_0, (1-\bar\alpha_t)\mathbf{I} \right)
\end{equation}
\begin{equation}
\label{eq:backwardinpaint}
    x^{\unknow}_{t-1} \sim \mathcal{N}\left( \mu_\theta(x_t,t), \Sigma_\theta(x_t,t) \right)
\end{equation}
\begin{equation}
\label{eq:repaint:combine}
    x_{t-1} = x^{\know}_{t-1} \odot m + x^{\unknow}_{t-1} \odot (1-m)
\end{equation}
\modelrepaint uses the known pixels as a noisy condition to guide the generation of the unknown pixels.
In the first inpainting step, inputs contain a high noise level and limited information about the condition. This leads to samples that deviate from the intended semantics of the condition, often resulting in artifacts. To address this, \modelrepaint introduces a resampling mechanism that harmonizes the two semantics by applying forward diffusion on the output $x_{t-1}$ back to $x_{t+j}$. The denoising and re-noising process involves executing the same diffusion step multiple times during the generation phase, sacrificing computational efficiency to achieve higher image quality.

\section{The Proposed \model Method}\label{section:method}

In \model, the model is conditioned on known and unknown regions using the time step $t$, already present in diffusion models.
Instead of using $x^{\know}_{t-1}$, $x_0$ is combined directly with $x_t$ without forward diffusion. This results in a faster diffusion process without changing the model architecture.

\subsection{Noise Modeling for Fast Inpainting}
\label{timemapours}
\begin{figure}
\centering
\begin{minipage}[t]{0.48\contentwidth}
  \includegraphics[width=1\linewidth]{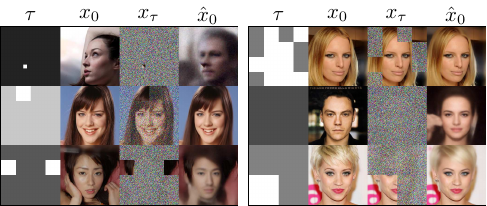}
  \captionof{figure}{\textbf{Illustration of the \model patch-wise training procedure.} Each region of the input $x_0$ receives a different level of noise controlled by $\tmap$. The network uses less noisy regions to reconstruct more noisy regions.}
\label{fig:method_train}
\end{minipage}%
\hfill
\begin{minipage}[t]{0.48\contentwidth}
  \includegraphics[width=1\linewidth]{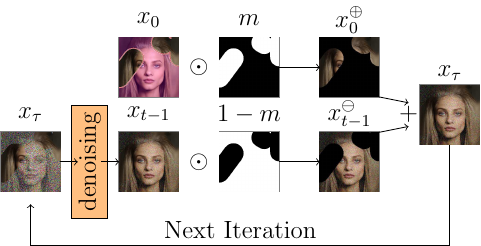}
  \captionof{figure}{\textbf{The conditional generation procedure}. \model modifies the standard denoising process to condition the diffusion model on the known region without noise while gradually denoising the generated region.}
\label{fig:method_generation}
\end{minipage}
\end{figure}

\textbf{Training Process:} The objective of \model is to enable the diffusion model to discern the information content of each input region. This allows the model to differentiate between conditioned regions and those needing to be painted.  By maintaining the known regions free of additional noise, \model preserves the maximum amount of information in these areas.
Using the time step $t$ allows for smooth and continuous modeling of information content. Regions from the known part of the image are given a $t$ value close to zero, indicating minimal noise. On the other hand, regions that need to be inpainted start with a $t$ value close to $T$, which progressively decreases during the generation process.
To do so, $t \in \{0,...,T\}$ is transformed from a scalar to a tensor $\tmap \in \{0,...,T\}^{h \times w}$.
Similarly, other variables are accommodated to perform a pixel-wise diffusion process. With one $t$ per pixel, $\alpha_t$ (resp. $\bar\alpha_t$, $\beta_t$) becomes $\alpha_\tmap$ (resp. $\bar\alpha_\tmap$, $\beta\tmap$).
This innovative modification can be integrated into most diffusion model training pipelines. \cref{fig:method_train} illustrates the intermediate images used for training.
During training, each input image pixel $x_{i,j}$ receives an amount of noise controlled by $\tmap_{i,j}$. The forward diffusion process is then applied to $x$ on a pixel-wise basis, as illustrated in \cref{fig:method_train} and can be formulated as:
\begin{equation}
    x_{\tmap} = \sqrt{\bar\alpha_{\tmap}}x_0 
        + 
        \sqrt{1-\bar\alpha_{\tmap}}\epsilon,
\end{equation}
in which $\epsilon \sim \mathcal{N}(\mathbf{0}, \mathbf{I})$, $\tmap \sim \phi_\text{train}$ and $\phi_\text{train}$ is a training strategy to sample different noise per pixel.
The diffusion network then predicts $\epsilon$, using less noisy regions to reconstruct more noisy regions by optimizing the loss:
\begin{equation}
    \mathcal{L} = \left\| \epsilon - \epsilon_\theta(x_{\tmap}, \tmap) \right\|^2_2.
\end{equation}

The final component of \model training is the strategy used for $\phi_\text{train}$. A random patch size and a proportion of patches are sampled to define a condition.
Based on this sampled proportion, the input image is then divided into known regions $x^{\know}$ and unknown regions $x^{\unknow}$ (with corresponding $t^{\know}=0$ and $t^{\unknow}=t$).
The possible patch sizes are defined as powers of two, i.e., ${ 2^i | i \in \mathbb{N}, 2^i \le w }$, up to the maximum size of the image.
The fraction of pixels designated as the known region is represented by a real value within the interval $[0, 1]$, ensuring that at least one patch remains in the unknown region. For example, as illustrated in \cref{fig:method_train} for a patch size of $128$, $25\%$ of the pixels are assigned to the known region.

\textbf{Generation Process:} Inpainting an image with \model involves sampling a time $t$ for the conditioning region $x^{\know}$ and a time $t$ for the inpainted region $x^{\unknow}$ using $\phi$.
Unless otherwise specified, we set $\phi^{\know}_t=0$ for the condition and $\phi^{\unknow}_t=t$ for the region to inpaint for training and generation.
For generation, the known region $x^{\know}$ is merged with the current unknown region $x^{\unknow}_t$, and one reverse step can be expressed as:
\begin{equation}
    x_{\tmap} = x^{\unknow}_t \odot (1-m) + x^{\know}_0 \odot m
\end{equation}
\begin{equation}
\label{eq:ours:combine}
    x^{\unknow}_{t-1} \sim \mathcal{N}\left( \mu_\theta(x_{\tmap},\tmap), \Sigma_\theta(x_{\tmap},\tmap) \right).
\end{equation}

This approach allows the use of input-known regions from the beginning of the diffusion process. Consequently, we can eliminate \modelrepaint's resampling mechanism, resulting in faster inpainting.
The general algorithm for inpainting an image with an arbitrary $\phi$ function is given by \cref{algo:model_generation}.
\newpage
\setlength{\intextsep}{3pt} 
\begin{wrapfigure}{L}{0.5\textwidth}
\begin{minipage}{0.5\textwidth}
\begin{algorithm}[H]
\caption{\model Generation Process.}
\label{algo:model_generation}

\begin{algorithmic}[1]
\REQUIRE $x^\know \sim q(x_0)$ a condition
\REQUIRE $m$ a condition mask
\REQUIRE $\phi_t$ giving the condition noise level for the known and unknown regions

\STATE $x_T \sim \mathcal{N}(\mathbf{0}, \mathbf{I})$

\FOR{$t = T,...,1$}
    \STATE $\epsilon \sim \mathcal{N}(0, \mathbf{I})$
    \STATE $z \sim \mathcal{N}(0, \mathbf{I})$ if $t>1$, else $z=0$
    
    \STATE $x^{\know}_{\phi_t^{\know}} =
        \sqrt{\bar\alpha_{\phi_t^{\know}}}x^\know
        + 
        \sqrt{1-\bar\alpha_{\phi_t^{\know}}}\epsilon$
    
    \STATE $\tmap = {\phi_t}^{\unknow} \odot (1-m) + {\phi_t}^{\know} \odot m$
    
    \STATE $x_{\tmap} = x^{\unknow}_{\phi_t^{\unknow}} \odot (1-m) + x^{\know}_{\phi_t^{\know}} \odot m$
    
    \STATE $x_{t-1} =
        \frac{1}{\sqrt{\bar\alpha_{\tmap}}}
        \left(
            x_\tmap - \frac{\beta_\tmap}{\sqrt{1 - \bar\alpha_\tmap}} \epsilon_\theta(x_\tmap, \tmap) 
        \right) + \sigma_\tmap z
    $
\ENDFOR
\RETURN $x_0$
    \end{algorithmic}
  \end{algorithm}
\end{minipage}
\end{wrapfigure}

\subsection{Time-aware Diffusion Architecture}

Diffusion models using U-Net architectures \citep{ronneberger2015u} can incorporate a time map $\tmap$ without requiring architectural modifications. This adaptation is achieved by applying pixel-wise time conditioning to each pixel in the feature maps.

Classical time conditioning first uses a position embedding layer to obtain a time embedding $\gamma \in \mathbb{R}^{d}$ from the time step $t$:
\begin{equation}
    \gamma = L\left[E(t) \times \sigma(L(E(t)))\right],
\end{equation}
where $L$ denotes linear layers, $\sigma$ is the sigmoid function, and $E$ is a sinusoidal timestep embedding.
Additionally, at each layer of the U-Net, the vector $\gamma$ is used to perform time conditioning with scale-shift normalization and can be written as:
\begin{equation}
    h^{l+1}_{i,j} = GN(h^{l})_{i,j} \times (1 + L_\text{scale}(\gamma)) + L_\text{shift}(\gamma),
\end{equation}
where $h^l \in \mathbb{R}^{c_l \times h_l \times w_l}$ are the current features for layer $l$, $GN$ is a group normalization layer, while $L_\text{scale}$ and $L_\text{shift}$ are linear layers that change the dimension of $\gamma$ from $d$ to $c_l$.
We apply the pixel-wise time conditioning across each spatial dimension $h^l$ by using $\tmap_{i,j}$ instead of $t$ as follows:
\begin{equation}
    \Gamma_{i,j} = L(E(D(\tmap)_{i, j}) \times \sigma(L(E(D(\tmap)_{i, j}))))
\end{equation}
\begin{equation}
    h^{l+1}_{i,j} = GN(h^{l})_{i,j} \times (1 + L_\text{scale}(\Gamma_{i,j})) + L_\text{shift}(\Gamma_{i,j}),
\end{equation}
where $\Gamma \in \mathbb{R}^{d \times h \times w}$ is the embedding of time $\tmap$, and $D$ is a downscaling operation function that rescales \tmap to $h_l \times w_l$\footnote{We use bilinear interpolation, but min-pool or other techniques could be used with a similar effect.}.

\section{Results and Discussion}
\label{partie:experiments}

This section empirically shows that \model:  
(a) produces high-quality inpainting with a clean condition on various mask sizes and shapes, on par or better than other inpainting models; 
(b) provides more efficient sampling, making it faster than other diffusion-based models without requiring a dedicated architecture; 
(c) generates diverse, high-quality images.
Details on the training masks and their corresponding ablation study are presented in \cref{ablation:trainingmask}, and additional qualitative results are provided in \cref{appendix:moreimages}.
In \cref{vsother}, \model is compared with state-of-the-art mask and image-conditioned inpainting models: the CNN-based \modellama \citep{suvorov2021resolutionrobust} and transformer-based \modelmat \citep{li2022mat}.
\vspace{2pt}

\begin{figure}[h]
    \centering
    \includegraphics[width=0.98\contentwidth]{assets/celebahq_grid1_2col.pdf}
    \caption{
    \textbf{Qualitative results:} \model against state-of-the-art inpainting methods on \celebahq. 
    Zoom in for better details.
    Additional examples can be found in \cref{appendix:moreimages}.
    }
    \label{fig:exp:celebahq}
\end{figure}

\begin{figure}[th]
    \centering
    \includegraphics[width=0.98\contentwidth]{assets/imagenet_grid_2col.pdf}
    \caption{
    \textbf{Qualitative results:} \model against state-of-the-art inpainting methods on \imagenet. 
    Additional examples can be found in \cref{appendix:moreimages}.
    }
    \label{fig:exp:imagenet}
\end{figure}

\begin{figure}[th]
    \centering
    \includegraphics[width=0.98\contentwidth]{assets/places_grid1_2col.pdf}
    \caption{
    \textbf{Qualitative results:} \model against state-of-the-art inpainting methods on \places. 
    Additional examples can be found in \cref{appendix:moreimages}.
    }
    \label{fig:exp:places}
\end{figure}

\subsection{Experimental Methodology} 
\label{implementationdetails}
\label{evaluationmetrics}

\noindent\textbf{Baselines:}
We evaluate \model against state-of-the-art diffusion-based inpainting methods in the pixel space: \modelrepaint \citep{lugmayr2022repaint}, which conditions the generative process using noisy inputs and synchronizes them with the output through resampling. \modelmcg \citep{chung2022improving}, which adds a correction term to keep the generation closer to the data manifold, and
\modelcopaint \citep{zhang2023towards} that utilizes Tweedie's formula for better generation.
Recent studies \citep{chung2022improving,zhang2023towards} have shown that \modelcopaint and \modelmcg are among the best-performing inpainting methods.
We denote the \modelrepaint-20 model as using 20 resampling steps, while \modelvanilla refers to the model with 1 resampling step matching the number of steps used by \model.
Furthermore, we conduct comparisons with latent diffusion models that have access to the complete context: \modelldm \citep{rombach2022highresolution} and \modelcontrolnet \citep{zhang2023adding}. 
Additionally, we compare against foundation model-based latent diffusion approaches, including Blended Latent Diffusion (\modelbld) \citep{avrahami2023blended}, \modelunipaint \citep{yang2023uni}, and \modelpowerpaint \citep{zhuang2025task}.

\noindent\textbf{Implementation Details:}
Our approach is validated using the \celebahq \citep{karras2018progressive} dataset, the \imagenet \citep{ILSVRC15} dataset, and the \places dataset \citep{places} at 256x256 resolution.
We modify the implementation of \citep{dhariwal2021diffusion}, maintaining all their hyperparameters. Training on \celebahq is conducted for approximately 150K steps with batch size 64 on 4 A100, for \imagenet and \places for about 200K steps with batch size 128 on 8 A100.
For baselines, we utilize existing code and pre-trained models when available.
For \imagenet, we train \modellama for 1M steps on batch size 5 using their implementation, and \modelmat for 300K steps on batch size 32 using their implementation.
Both \modelldm and \modelcontrolnet are trained using computational resources equivalent to \model. For \modelldm, the encoded masked image and the downsampled mask provide additional context during the sampling process.
For \modelcontrolnet, the encoded masked image alone provides additional context.

\noindent\textbf{Evaluation Metrics:}
Image quality is evaluated using established metrics from the inpainting literature: the Learned Perceptual Image Patch Similarity \citep{zhang2018unreasonable} (\lpips), the Structural Similarity Index Measure \citep{ssim} (\ssim), and the Kernel Inception Distance \citep{binkowski2018demystifying} (\kid) (using the TorchMetrics \citep{torchmetric} implementation).
The number of diffusion steps (\nbsteps) and the mean time to inpaint an image (Runtime) are used to evaluate the computational efficiency of \model.
For evaluation, we use 2,824 images from the \celebahq test set, 5,000 images from \imagenet, and 2,000 images from \places.

\subsection{Comparison with Diffusion-Based Models}\label{results:diffusion}

\noindent\textbf{\wide and \narrow masks}
In the standard image inpainting scenario, \model is compared using \wide and \narrow masks following \modellama \citep{suvorov2021resolutionrobust} protocol.
\Cref{table:exp:celebahq_imagenet:diffusion,table:exp:kid_table} shows that \model consistently outperforms other diffusion-based models, improving by 20\% \modelrepaint-20 \lpips's on the \wide mask on \celebahq, \imagenet and \places, and by 30\% on the \narrow masks.
\modelmcg lacks global consistency, resulting in significant artifacts on \wide masks, as seen in \cref{fig:exp:celebahq,fig:exp:imagenet} where it produces eyes of different colors, strange textures, and inpainting artifacts. \modelrepaint produces high-quality images at the cost of significant inference time. Our approach can produce high-quality images while requiring much less processing time.
When processing small masks, \modelldm occasionally produces minor artifacts, as evident in the top rows of \cref{fig:exp:celebahq,fig:exp:imagenet}.
These artifacts are more pronounced in \modelcontrolnet, which lacks mask information, resulting in inconsistencies between known and unknown regions.
Among foundation-based models, \modelbld demonstrates superior performance across all datasets, while \modelunipaint and \modelpowerpaint show inferior results on \celebahq but achieving better performance on \imagenet and \places.
This performance variation may be attributed to \modelbld's use of the provided LDM baselines (our early experiments with Stable Diffusion yielded less favorable results), whereas both \modelunipaint and \modelpowerpaint utilize Stable Diffusion as their foundation.

\noindent\textbf{\twotime and \altline masks}
In the \twotime setting, every other pixel is removed from the image, while the \altline setting removes every other line.
All pixel-based baselines achieve low \lpips scores and produce high-quality images for both types of masks (see \cref{fig:exp:celebahq,fig:exp:imagenet,fig:exp:places}).
In contrast, latent-based models prove inadequate for this task due to their downsampling of inpainting masks, which results in critical information loss.
\model outperforms all considered baselines on \celebahq and comes a close second to \modelcopaint on \imagenet and \places, as shown in \cref{table:exp:celebahq_imagenet:diffusion,table:exp:kid_table}.
The third-best performing model is \modelrepaint. Compared to \modelrepaint, \model improves \lpips by 115\% in the \twotime setting and by 69\% in the \altline setting.

\noindent\textbf{\half and \expand masks}
The \half setting removes the right part of the images, and the \expand setting keeps only the central 64$\times$64 part of a 256$\times$256 pixel image.
Both \lpips and \ssim metrics are less suitable when a significant portion of the image is missing, as they rely on a single ground truth. This penalizes methods that generate realistic images with semantics different from the ground truth \citep{lugmayr2022repaint}.
In this case, the \kid, which measures the distance between distributions, is more reliable for assessing image quality.
Applying \half and \expand masks is particularly challenging, as they remove substantial portions of the images. task where an important part of the images is removed. This complexity is shown both visually in \cref{fig:exp:celebahq,fig:exp:imagenet,fig:exp:places} and by the quantitative results in \cref{table:exp:celebahq_imagenet:diffusion,table:exp:kid_table}.
Our model performs best on both the \half and \expand masks across all datasets (except for \celebahq where it comes close to \modelldm), as measured by the \kid, while significantly reducing the inference time.
\Cref{fig:exp:celebahq} shows that \model can produce high-quality images in this challenging setting where \modelrepaint lacks global consistency.
On \imagenet and \places (see \cref{fig:exp:imagenet,fig:exp:places}), we observe that \modelmcg and \modelcopaint often mirror the images from the \half mask, resulting in symmetrical outputs.
Similar behavior is observed on \expand masks, with significant texture blending.
\modelcontrolnet exhibits limited generalization capability when handling very large masks, likely due to the absence of mask information, whereas \modelldm demonstrates robust performance in these scenarios.
Our analysis reveals that foundation-based models yield inferior qualitative results on \celebahq and \places datasets. However, they show improved performance on \imagenet, where class names provide supplementary textual information, which is particularly beneficial in scenarios with limited contextual information.

\begin{table*}
    \centering
    \caption{
    \textbf{Quantitative results:} \lpips and \ssim evaluation of diffusion models for inpainting on the \celebahq, \imagenet and \places datasets.}
    \includegraphics[width=1\contentwidth]{tables/celebahq_and_imagenet_diffusion.tex}
    \label{table:exp:celebahq_imagenet:diffusion}
\end{table*}

\begin{table*}[!t]
    \centering
    \caption{
    \textbf{Quantitative results:} \kid evaluation of diffusion models for inpainting on the \celebahq, \imagenet and \places datasets.}
    \includegraphics[width=0.8\contentwidth]{tables/kid_table.tex}
    \label{table:exp:kid_table}
\end{table*}

\subsection{Quality vs. Efficiency}\label{results:speed}

\begin{wraptable}{R}{0.5\textwidth}
    \centering
    \caption{Inpainting speed for different diffusion model in the pixel space.}
    \includegraphics[]{tables/celebahq_narrow_runtime_table.tex}
    \label{table:exp:celebahq_runtime}
\end{wraptable} 

We compare the time efficiency of different diffusion approaches working in the pixel space by computing the average time to sample 100 images consecutively on a single V100, and the results are reported in \cref{table:exp:celebahq_runtime}.
State-of-the-art approaches require over $6\times$ longer to sample compared to \model. The increased time in \modelrepaint is due to the resampling mechanism needed to synchronize condition and generation, while \modelmcg requires an additional backward pass and more steps to optimize the image.
In contrast, our model reduces inference time by trading fine-tuning costs, enabling faster generation of high-quality images than other diffusion-based inpainting models.

\begin{figure}[h]
     \centering
    \includegraphics[width=0.90\contentwidth]{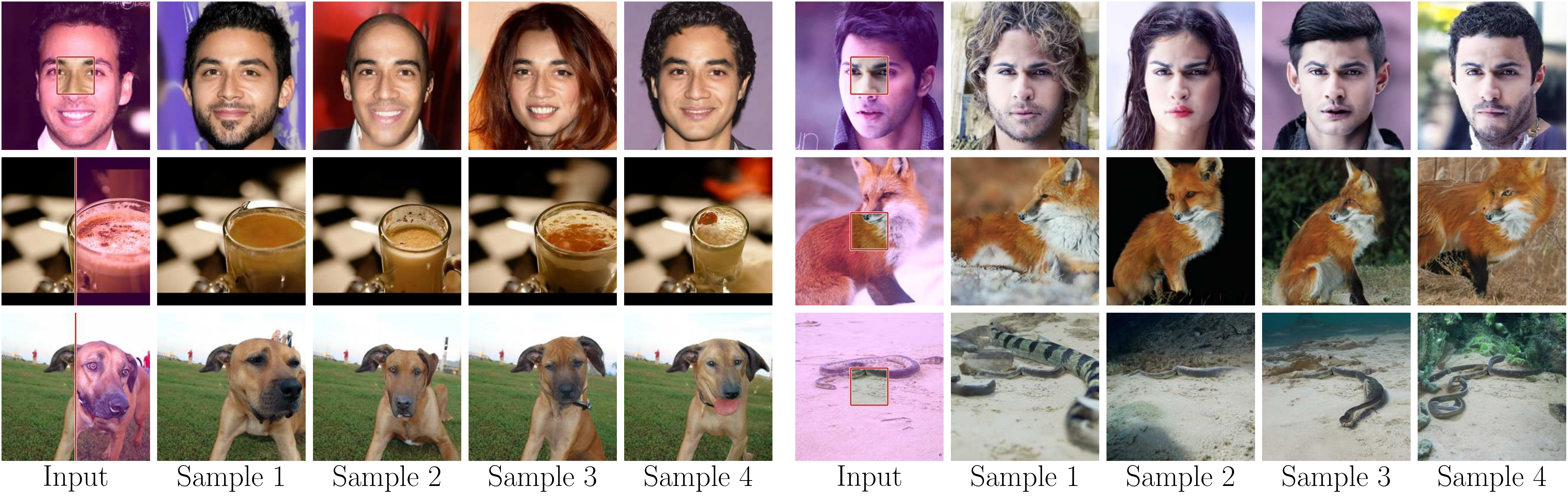}
    \caption{Examples of diverse generations using \model on the CelebA and ImageNet, with the same input image and different initial noise. Additional examples are available in \cref{appendix:moreimages}.}
    \label{fig:exp:diversity}
\end{figure}

\subsection{Diversity of Generated Images}
While \model performs fast and high-quality inpainting, we must ask whether this comes at the cost of diversity. To evaluate this, we compute the Diversity Score \citep{lugmayr2021ntire} by generating 10 different images for 100 inputs. The quantitative results reported in \cref{table:exp:celebahq_diversity}.
The most diverse model is \modelvanilla, which also has a high \lpips score. In contrast, \model achieves a high Diversity Score across most masks while consistently producing high-quality images (see \cref{fig:exp:diversity}) with lower \lpips scores.

\vspace{9pt}
\begin{table}[h]
    \centering
    \caption{Diversity Score on \celebahq with 10 random generations across 100 images.}
    \includegraphics[width=0.75\contentwidth]{tables/celebahq_diversity.tex}
    \label{table:exp:celebahq_diversity}
\end{table}

\section{Conclusions}
In this paper, we introduce \modelfull (\model), a method that accelerates inpainting by modeling multiple noise levels through time conditioning in the diffusion process. Unlike other diffusion-based models, \model does not require any special architecture for inpainting, and generates high-quality, diverse images more quickly. This efficiency makes \model highly practical for real-world applications and usable for resource-constrained devices.
\section*{Acknowledgments}
This work was financially supported by the ANR Labcom LLisa ANR-20-LCV1-0009.
We thank CRIANN, who provided us with the computation resources necessary for our experiments.
This work was performed using HPC resources from GENCI–IDRIS (Grant 2024-AD011013862R1).
We thank the Natural Sciences and Engineering Research Council of Canada (NSERC) and the Digital Research Alliance of Canada.

\bibliography{iclr2025_conference}

\begin{thebibliography}{42}
\providecommand{\natexlab}[1]{#1}
\providecommand{\url}[1]{\texttt{#1}}
\expandafter\ifx\csname urlstyle\endcsname\relax
  \providecommand{\doi}[1]{doi: #1}\else
  \providecommand{\doi}{doi: \begingroup \urlstyle{rm}\Url}\fi

\bibitem[Avrahami et~al.(2023)Avrahami, Fried, and Lischinski]{avrahami2023blended}
Omri Avrahami, Ohad Fried, and Dani Lischinski.
\newblock Blended latent diffusion.
\newblock \emph{ACM transactions on graphics (TOG)}, 42\penalty0 (4):\penalty0 1--11, 2023.

\bibitem[Bi{\'n}kowski et~al.(2018)Bi{\'n}kowski, Sutherland, Arbel, and Gretton]{binkowski2018demystifying}
Miko{\l}aj Bi{\'n}kowski, Danica~J Sutherland, Michael Arbel, and Arthur Gretton.
\newblock Demystifying mmd gans.
\newblock \emph{arXiv preprint arXiv:1801.01401}, 2018.

\bibitem[Chung et~al.(2022)Chung, Sim, Ryu, and Ye]{chung2022improving}
Hyungjin Chung, Byeongsu Sim, Dohoon Ryu, and Jong~Chul Ye.
\newblock Improving diffusion models for inverse problems using manifold constraints.
\newblock \emph{Advances in Neural Information Processing Systems}, 35:\penalty0 25683--25696, 2022.

\bibitem[Criminisi et~al.(2004)Criminisi, Perez, and Toyama]{1323101}
A.~Criminisi, P.~Perez, and K.~Toyama.
\newblock Region filling and object removal by exemplar-based image inpainting.
\newblock \emph{IEEE Transactions on Image Processing}, 13\penalty0 (9):\penalty0 1200--1212, 2004.

\bibitem[Dauphin et~al.(2017)Dauphin, Fan, Auli, and Grangier]{dauphin2017language}
Yann~N Dauphin, Angela Fan, Michael Auli, and David Grangier.
\newblock Language modeling with gated convolutional networks.
\newblock In \emph{International conference on machine learning}, pp.\  933--941, 2017.

\bibitem[Dhariwal \& Nichol(2021)Dhariwal and Nichol]{dhariwal2021diffusion}
Prafulla Dhariwal and Alexander Nichol.
\newblock Diffusion models beat gans on image synthesis.
\newblock \emph{Advances in neural information processing systems}, 34:\penalty0 8780--8794, 2021.

\bibitem[Goodfellow et~al.(2020)Goodfellow, Pouget-Abadie, Mirza, Xu, Warde-Farley, Ozair, Courville, and Bengio]{goodfellow2014generative}
Ian Goodfellow, Jean Pouget-Abadie, Mehdi Mirza, Bing Xu, David Warde-Farley, Sherjil Ozair, Aaron Courville, and Yoshua Bengio.
\newblock Generative adversarial networks.
\newblock \emph{Communications of the ACM}, 63\penalty0 (11):\penalty0 139--144, 2020.

\bibitem[Grossauer(2004)]{10.1007/978-3-540-24671-8_17}
Harald Grossauer.
\newblock A combined pde and texture synthesis approach to inpainting.
\newblock In Tom{\'a}s Pajdla and Ji{\v{r}}{\'i} Matas (eds.), \emph{Proceedings of the European conference on computer vision}, pp.\  214--224, 2004.

\bibitem[Ho et~al.(2020)Ho, Jain, and Abbeel]{ho2020denoising}
Jonathan Ho, Ajay Jain, and Pieter Abbeel.
\newblock Denoising diffusion probabilistic models.
\newblock \emph{Advances in neural information processing systems}, 33:\penalty0 6840--6851, 2020.

\bibitem[Iizuka et~al.(2017)Iizuka, Simo-Serra, and Ishikawa]{iizuka2017globally}
Satoshi Iizuka, Edgar Simo-Serra, and Hiroshi Ishikawa.
\newblock Globally and locally consistent image completion.
\newblock \emph{ACM Transactions on Graphics (ToG)}, 36\penalty0 (4):\penalty0 1--14, 2017.

\bibitem[Karras et~al.(2018)Karras, Aila, Laine, and Lehtinen]{karras2018progressive}
Tero Karras, Timo Aila, Samuli Laine, and Jaakko Lehtinen.
\newblock Progressive growing of gans for improved quality, stability, and variation.
\newblock In \emph{International Conference on Learning Representations}, 2018.

\bibitem[Karras et~al.(2019)Karras, Laine, and Aila]{karras2019stylebased}
Tero Karras, Samuli Laine, and Timo Aila.
\newblock A style-based generator architecture for generative adversarial networks.
\newblock In \emph{Proceedings of the IEEE/CVF conference on computer vision and pattern recognition}, pp.\  4401--4410, 2019.

\bibitem[Kingma \& Welling(2013)Kingma and Welling]{kingma2022autoencoding}
Diederik~P Kingma and Max Welling.
\newblock Auto-encoding variational bayes.
\newblock \emph{arXiv preprint arXiv:1312.6114}, 2013.

\bibitem[Levin \& Fried(2023)Levin and Fried]{levin2023differential}
Eran Levin and Ohad Fried.
\newblock Differential diffusion: Giving each pixel its strength.
\newblock \emph{arXiv preprint arXiv:2306.00950}, 2023.

\bibitem[Li et~al.(2022)Li, Lin, Zhou, Qi, Wang, and Jia]{li2022mat}
Wenbo Li, Zhe Lin, Kun Zhou, Lu~Qi, Yi~Wang, and Jiaya Jia.
\newblock Mat: Mask-aware transformer for large hole image inpainting.
\newblock In \emph{Proceedings of the IEEE/CVF conference on computer vision and pattern recognition}, pp.\  10758--10768, 2022.

\bibitem[Li et~al.(2023)Li, Yu, Zhou, Song, and Lin]{li2023image}
Wenbo Li, Xin Yu, Kun Zhou, Yibing Song, and Zhe Lin.
\newblock Image inpainting via iteratively decoupled probabilistic modeling.
\newblock In \emph{The Twelfth International Conference on Learning Representations}, 2023.

\bibitem[Liu et~al.(2018)Liu, Reda, Shih, Wang, Tao, and Catanzaro]{liu2018image}
Guilin Liu, Fitsum~A Reda, Kevin~J Shih, Ting-Chun Wang, Andrew Tao, and Bryan Catanzaro.
\newblock Image inpainting for irregular holes using partial convolutions.
\newblock In \emph{Proceedings of the European conference on computer vision}, pp.\  85--100, 2018.

\bibitem[Lugmayr et~al.(2021)Lugmayr, Danelljan, and Timofte]{lugmayr2021ntire}
Andreas Lugmayr, Martin Danelljan, and Radu Timofte.
\newblock Ntire 2021 learning the super-resolution space challenge.
\newblock In \emph{Proceedings of the IEEE/CVF Conference on Computer Vision and Pattern Recognition}, pp.\  596--612, 2021.

\bibitem[Lugmayr et~al.(2022)Lugmayr, Danelljan, Romero, Yu, Timofte, and Van~Gool]{lugmayr2022repaint}
Andreas Lugmayr, Martin Danelljan, Andres Romero, Fisher Yu, Radu Timofte, and Luc Van~Gool.
\newblock Repaint: Inpainting using denoising diffusion probabilistic models.
\newblock In \emph{Proceedings of the IEEE/CVF conference on computer vision and pattern recognition}, pp.\  11461--11471, 2022.

\bibitem[Mirza \& Osindero(2014)Mirza and Osindero]{mirza2014conditional}
Mehdi Mirza and Simon Osindero.
\newblock Conditional generative adversarial nets, 2014.

\bibitem[{Nicki Skafte Detlefsen} et~al.(2022){Nicki Skafte Detlefsen}, {Jiri Borovec}, {Justus Schock}, {Ananya Harsh}, {Teddy Koker}, {Luca Di Liello}, {Daniel Stancl}, {Changsheng Quan}, {Maxim Grechkin}, and {William Falcon}]{torchmetric}
{Nicki Skafte Detlefsen}, {Jiri Borovec}, {Justus Schock}, {Ananya Harsh}, {Teddy Koker}, {Luca Di Liello}, {Daniel Stancl}, {Changsheng Quan}, {Maxim Grechkin}, and {William Falcon}.
\newblock {TorchMetrics - Measuring Reproducibility in PyTorch}, February 2022.
\newblock URL \url{https://github.com/Lightning-AI/torchmetrics}.

\bibitem[Pan et~al.(2021)Pan, Zhan, Dai, Lin, Loy, and Luo]{pan2020exploiting}
Xingang Pan, Xiaohang Zhan, Bo~Dai, Dahua Lin, Chen~Change Loy, and Ping Luo.
\newblock Exploiting deep generative prior for versatile image restoration and manipulation.
\newblock \emph{IEEE Transactions on Pattern Analysis and Machine Intelligence}, 44\penalty0 (11):\penalty0 7474--7489, 2021.

\bibitem[Pathak et~al.(2016)Pathak, Krahenbuhl, Donahue, Darrell, and Efros]{pathak2016context}
Deepak Pathak, Philipp Krahenbuhl, Jeff Donahue, Trevor Darrell, and Alexei~A Efros.
\newblock Context encoders: Feature learning by inpainting.
\newblock In \emph{Proceedings of the IEEE conference on computer vision and pattern recognition}, pp.\  2536--2544, 2016.

\bibitem[Richardson et~al.(2021)Richardson, Alaluf, Patashnik, Nitzan, Azar, Shapiro, and Cohen-Or]{richardson2021encoding}
Elad Richardson, Yuval Alaluf, Or~Patashnik, Yotam Nitzan, Yaniv Azar, Stav Shapiro, and Daniel Cohen-Or.
\newblock Encoding in style: a stylegan encoder for image-to-image translation.
\newblock In \emph{Proceedings of the IEEE/CVF conference on computer vision and pattern recognition}, pp.\  2287--2296, 2021.

\bibitem[Rombach et~al.(2022)Rombach, Blattmann, Lorenz, Esser, and Ommer]{rombach2022highresolution}
Robin Rombach, Andreas Blattmann, Dominik Lorenz, Patrick Esser, and Bj{\"o}rn Ommer.
\newblock High-resolution image synthesis with latent diffusion models.
\newblock In \emph{Proceedings of the IEEE/CVF conference on computer vision and pattern recognition}, pp.\  10684--10695, 2022.

\bibitem[Ronneberger et~al.(2015)Ronneberger, Fischer, and Brox]{ronneberger2015u}
Olaf Ronneberger, Philipp Fischer, and Thomas Brox.
\newblock U-net: Convolutional networks for biomedical image segmentation.
\newblock In \emph{International Conference Medical image computing and computer-assisted intervention}, pp.\  234--241, 2015.

\bibitem[Russakovsky et~al.(2015)Russakovsky, Deng, Su, Krause, Satheesh, Ma, Huang, Karpathy, Khosla, Bernstein, Berg, and Fei-Fei]{ILSVRC15}
Olga Russakovsky, Jia Deng, Hao Su, Jonathan Krause, Sanjeev Satheesh, Sean Ma, Zhiheng Huang, Andrej Karpathy, Aditya Khosla, Michael Bernstein, Alexander~C. Berg, and Li~Fei-Fei.
\newblock {ImageNet Large Scale Visual Recognition Challenge}.
\newblock \emph{International Journal of Computer Vision}, 115\penalty0 (3):\penalty0 211--252, 2015.

\bibitem[Sohl-Dickstein et~al.(2015)Sohl-Dickstein, Weiss, Maheswaranathan, and Ganguli]{sohldickstein2015deep}
Jascha Sohl-Dickstein, Eric Weiss, Niru Maheswaranathan, and Surya Ganguli.
\newblock Deep unsupervised learning using nonequilibrium thermodynamics.
\newblock In \emph{International conference on machine learning}, pp.\  2256--2265, 2015.

\bibitem[Song et~al.(2020{\natexlab{a}})Song, Meng, and Ermon]{song2022denoising}
Jiaming Song, Chenlin Meng, and Stefano Ermon.
\newblock Denoising diffusion implicit models.
\newblock In \emph{International Conference on Learning Representations}, 2020{\natexlab{a}}.

\bibitem[Song et~al.(2020{\natexlab{b}})Song, Sohl-Dickstein, Kingma, Kumar, Ermon, and Poole]{song2021scorebased}
Yang Song, Jascha Sohl-Dickstein, Diederik~P Kingma, Abhishek Kumar, Stefano Ermon, and Ben Poole.
\newblock Score-based generative modeling through stochastic differential equations.
\newblock In \emph{International Conference on Learning Representations}, 2020{\natexlab{b}}.

\bibitem[Suvorov et~al.(2022)Suvorov, Logacheva, Mashikhin, Remizova, Ashukha, Silvestrov, Kong, Goka, Park, and Lempitsky]{suvorov2021resolutionrobust}
Roman Suvorov, Elizaveta Logacheva, Anton Mashikhin, Anastasia Remizova, Arsenii Ashukha, Aleksei Silvestrov, Naejin Kong, Harshith Goka, Kiwoong Park, and Victor Lempitsky.
\newblock Resolution-robust large mask inpainting with fourier convolutions.
\newblock In \emph{Proceedings of the IEEE/CVF winter conference on applications of computer vision}, pp.\  2149--2159, 2022.

\bibitem[Wang et~al.(2004)Wang, Bovik, Sheikh, and Simoncelli]{ssim}
Zhou Wang, A.C. Bovik, H.R. Sheikh, and E.P. Simoncelli.
\newblock Image quality assessment: from error visibility to structural similarity.
\newblock \emph{IEEE Transactions on Image Processing}, 13\penalty0 (4):\penalty0 600--612, 2004.

\bibitem[Yang et~al.(2023)Yang, Chen, and Liao]{yang2023uni}
Shiyuan Yang, Xiaodong Chen, and Jing Liao.
\newblock Uni-paint: A unified framework for multimodal image inpainting with pretrained diffusion model.
\newblock In \emph{Proceedings of the 31st ACM International Conference on Multimedia}, pp.\  3190--3199, 2023.

\bibitem[Yu \& Koltun(2015)Yu and Koltun]{yu2015multi}
Fisher Yu and Vladlen Koltun.
\newblock Multi-scale context aggregation by dilated convolutions.
\newblock \emph{arXiv preprint arXiv:1511.07122}, 2015.

\bibitem[Yu et~al.(2018)Yu, Lin, Yang, Shen, Lu, and Huang]{yu2018generative}
Jiahui Yu, Zhe Lin, Jimei Yang, Xiaohui Shen, Xin Lu, and Thomas~S Huang.
\newblock Generative image inpainting with contextual attention.
\newblock In \emph{Proceedings of the IEEE conference on computer vision and pattern recognition}, pp.\  5505--5514, 2018.

\bibitem[Yu et~al.(2019)Yu, Lin, Yang, Shen, Lu, and Huang]{yu2019free}
Jiahui Yu, Zhe Lin, Jimei Yang, Xiaohui Shen, Xin Lu, and Thomas~S Huang.
\newblock Free-form image inpainting with gated convolution.
\newblock In \emph{Proceedings of the IEEE/CVF international conference on computer vision}, pp.\  4471--4480, 2019.

\bibitem[Zeng et~al.(2019)Zeng, Fu, Chao, and Guo]{zeng2019learning}
Yanhong Zeng, Jianlong Fu, Hongyang Chao, and Baining Guo.
\newblock Learning pyramid-context encoder network for high-quality image inpainting.
\newblock In \emph{Proceedings of the IEEE/CVF conference on computer vision and pattern recognition}, pp.\  1486--1494, 2019.

\bibitem[Zhang et~al.(2023{\natexlab{a}})Zhang, Ji, Zhang, Yu, Jaakkola, and Chang]{zhang2023towards}
Guanhua Zhang, Jiabao Ji, Yang Zhang, Mo~Yu, Tommi Jaakkola, and Shiyu Chang.
\newblock Towards coherent image inpainting using denoising diffusion implicit models.
\newblock In \emph{International Conference on Machine Learning}, pp.\  41164--41193. PMLR, 2023{\natexlab{a}}.

\bibitem[Zhang et~al.(2023{\natexlab{b}})Zhang, Rao, and Agrawala]{zhang2023adding}
Lvmin Zhang, Anyi Rao, and Maneesh Agrawala.
\newblock Adding conditional control to text-to-image diffusion models.
\newblock In \emph{Proceedings of the IEEE/CVF International Conference on Computer Vision}, pp.\  3836--3847, 2023{\natexlab{b}}.

\bibitem[Zhang et~al.(2018)Zhang, Isola, Efros, Shechtman, and Wang]{zhang2018unreasonable}
Richard Zhang, Phillip Isola, Alexei~A Efros, Eli Shechtman, and Oliver Wang.
\newblock The unreasonable effectiveness of deep features as a perceptual metric.
\newblock In \emph{Proceedings of the IEEE conference on computer vision and pattern recognition}, pp.\  586--595, 2018.

\bibitem[Zhou et~al.(2018)Zhou, Lapedriza, Khosla, Oliva, and Torralba]{places}
Bolei Zhou, Agata Lapedriza, Aditya Khosla, Aude Oliva, and Antonio Torralba.
\newblock Places: A 10 million image database for scene recognition.
\newblock \emph{IEEE Transactions on Pattern Analysis and Machine Intelligence}, 40\penalty0 (6):\penalty0 1452--1464, 2018.
\newblock \doi{10.1109/TPAMI.2017.2723009}.

\bibitem[Zhuang et~al.(2025)Zhuang, Zeng, Liu, Yuan, and Chen]{zhuang2025task}
Junhao Zhuang, Yanhong Zeng, Wenran Liu, Chun Yuan, and Kai Chen.
\newblock A task is worth one word: Learning with task prompts for high-quality versatile image inpainting.
\newblock In \emph{European Conference on Computer Vision}, pp.\  195--211. Springer, 2025.

\end{thebibliography}
\bibliographystyle{iclr2025_conference}

\appendix
\section{Ablations: Training Mask}
\label{ablation:trainingmask}
We examine the contribution of masks during training on the \celebahq dataset using our mask strategy (\mto) described in \cref{timemapours}, \modellama masks (\mtl), and a random mix of the two (\mtol). Unless stated otherwise, all results are reported with \mtol.
\Cref{table:exp:celebahq_masksused} shows that using \mtl over \mto reduces the error for every test mask except \twotime and \altline, which is explained by the more complicated design of \mtl, which produces train masks closer to the \wide and \narrow test masks and of more diverse shapes.
Notably, using \mtol masks still allows for very low \twotime and \altline errors compared to \modellama errors on the same masks in \cref{table:exp:celebahq_imagenet:gan_transformer}.
Using \mtol allows the benefits of \mto masks to be retrained while having low \twotime and \altline errors.

\begin{table*}[th]
\centering
\caption{
\textbf{Ablation study for the types of training time map.}
The metrics show how the use of \mto focuses more on very fine inpainting masks.
The distribution of \mtl masks is closer to larger inpainting masks, such as \wide.
Combining the two allows for strong performance across the range of test masks considered.
}
\includegraphics[width=1\contentwidth]{tables/celebahq_mask_ablation.tex}
\label{table:exp:celebahq_masksused}
\end{table*}


\section{Image Quality and Diffusion Steps}
\label{trainingstepvslpips}
\begin{figure}[tb]
\centering
\includegraphics[width=\contentwidth]
{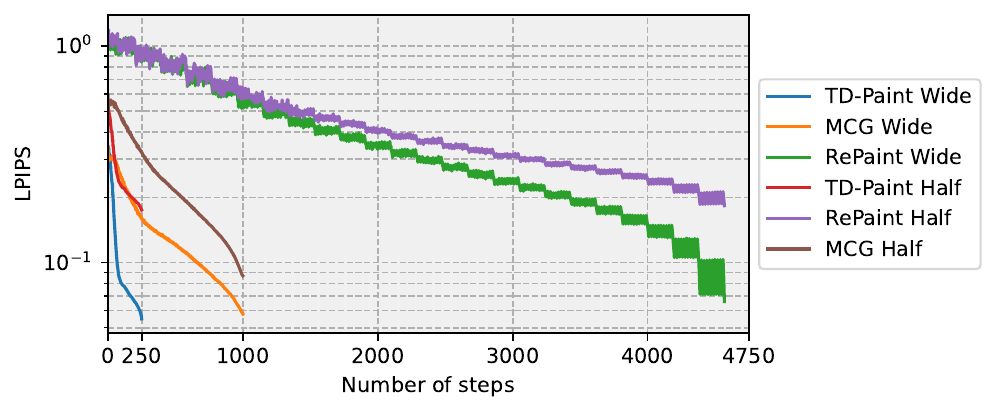}
\caption{Image quality at different time step for 100 images on \celebahq dataset for \wide and \half masks.}
\label{fig:face:lower_supervision}
\end{figure}

\begin{figure}[tb]
\centering
\includegraphics[width=\contentwidth]
{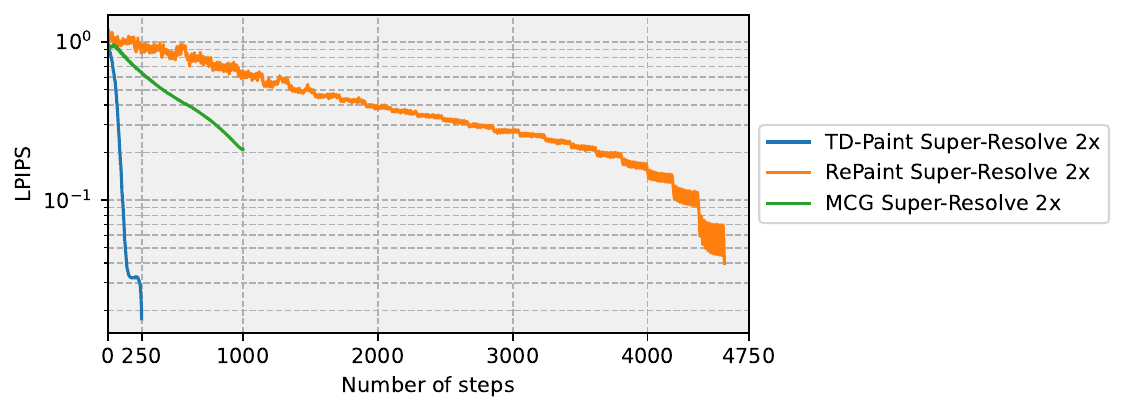}
\caption{Image quality at different time step for 100 images on \celebahq dataset for \twotime mask.}
\label{fig:face:lower_supervision:nn2}
\end{figure}

We compare the \lpips metrics over the diffusion step for \model, \modelrepaint and \modelmcg in \cref{fig:face:lower_supervision,fig:face:lower_supervision:nn2} averaged over 100 images from the \celebahq dataset.
\model can produce high-quality images in a fraction of the steps required by \modelrepaint because it takes advantage of the state since the early step of the diffusion process.

\section{Comparison with CNN- and Transformer-Based Models}
\label{vsother}

We compare \model with \modellama (CNN-based) and \modelmat (transformer-based) models in \cref{table:exp:celebahq_imagenet:gan_transformer,table:exp:kid_table:others}.

\begin{figure}[th]
    \centering
    \includegraphics[width=0.98\contentwidth]{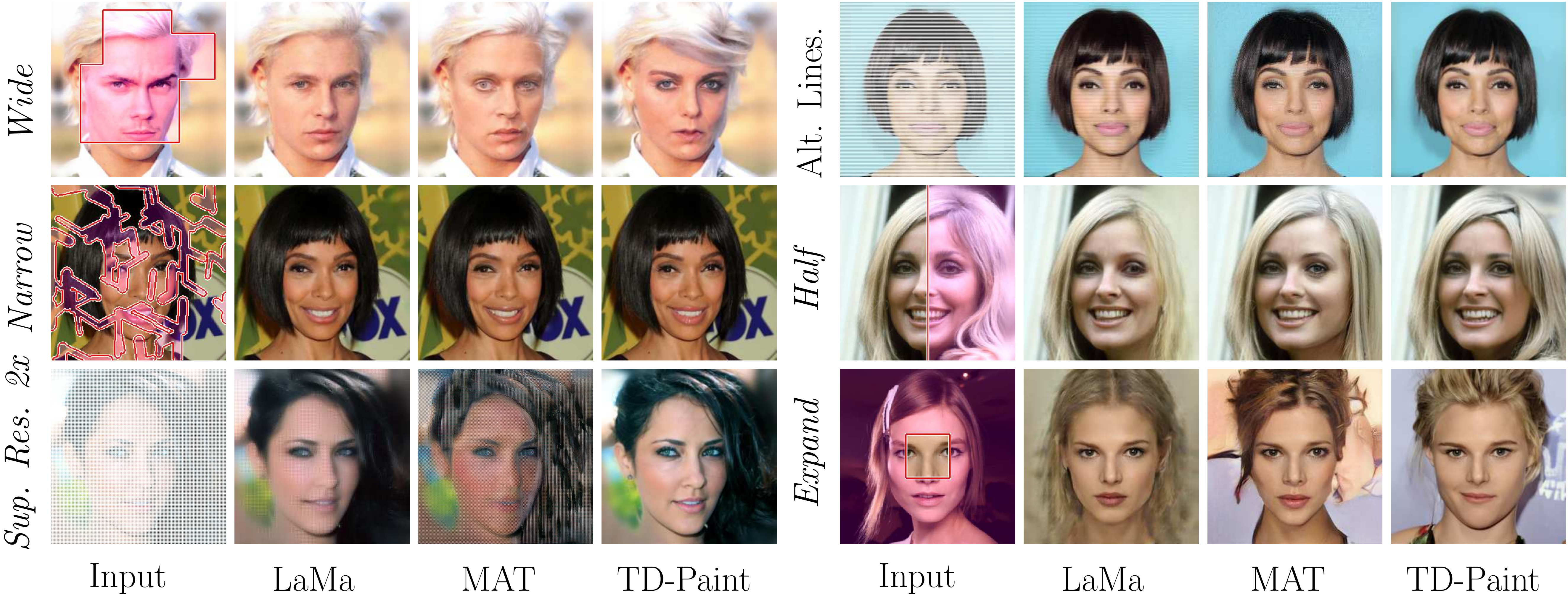}
    \caption{
    \textbf{Qualitative results:} \model against state-of-the-art inpainting CNN- and transformer-based models on \celebahq. 
    Zoom in for better details.
    Additional examples can be found in \cref{appendix:moreimages}.
    }
    \label{fig:exp:celebahq:others}
\end{figure}

\noindent\textbf{\wide and \narrow masks}
Our approach closely matches the performance of \modellama and \modelmat in the \wide setting on \celebahq and even surpasses them in the \narrow setting on \celebahq, as well as in both the \wide and \narrow settings on \imagenet. As shown in \cref{fig:exp:celebahq:others}, \modellama tends to generate pupils of different sizes when one eye is hidden in the \wide and different eye colors in the \narrow settings.

\begin{figure}[th]
    \centering
    \includegraphics[width=0.98\contentwidth]{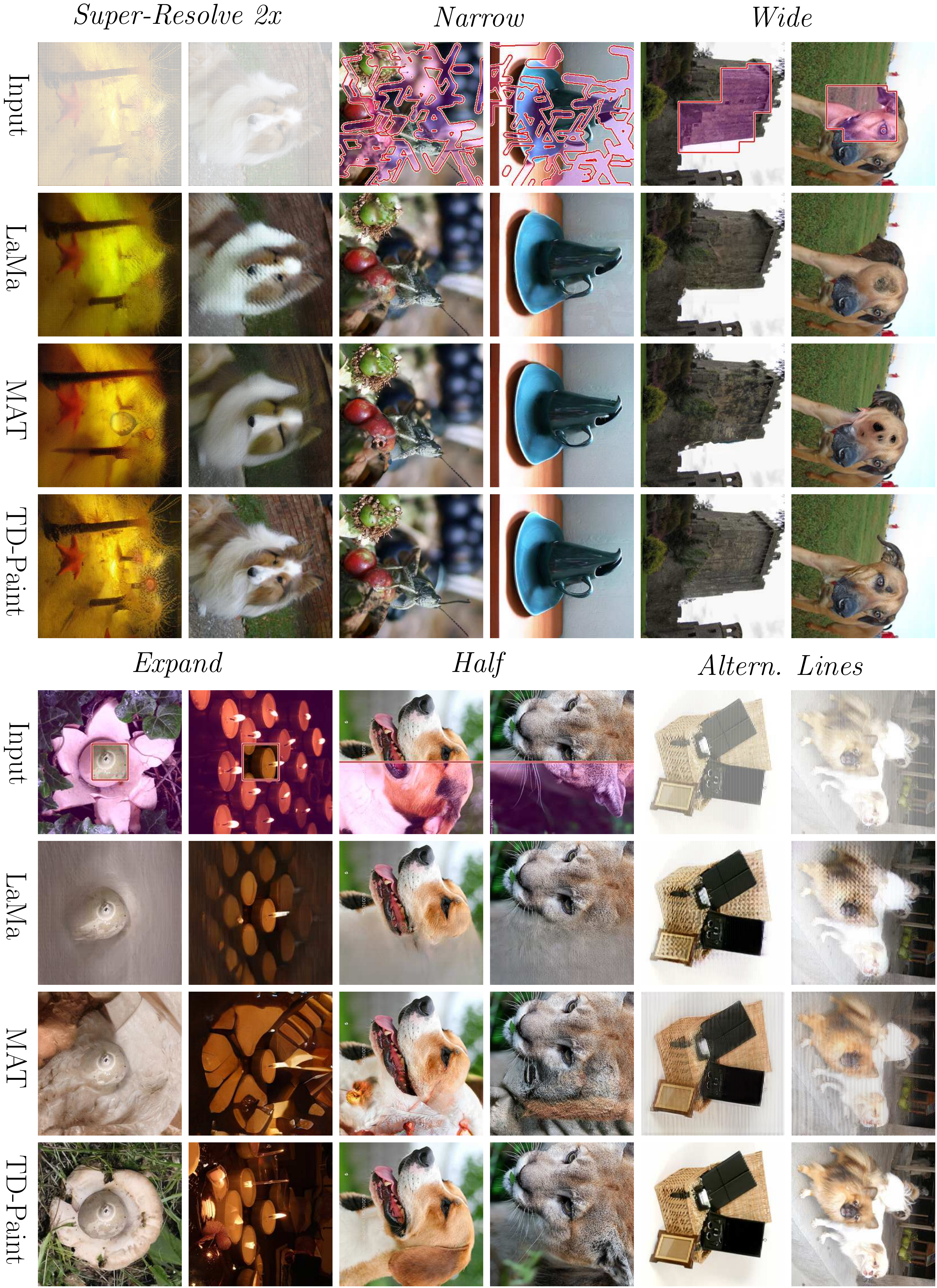}
    \caption{
    \textbf{Qualitative results:} \model against state-of-the-art inpainting CNN- and transformer-based models on \imagenet. 
    Additional examples can be found in \cref{appendix:moreimages}.
    }
    \label{fig:exp:imagenet:others}
\end{figure}

\begin{figure}[th]
    \centering
    \includegraphics[width=0.98\contentwidth]{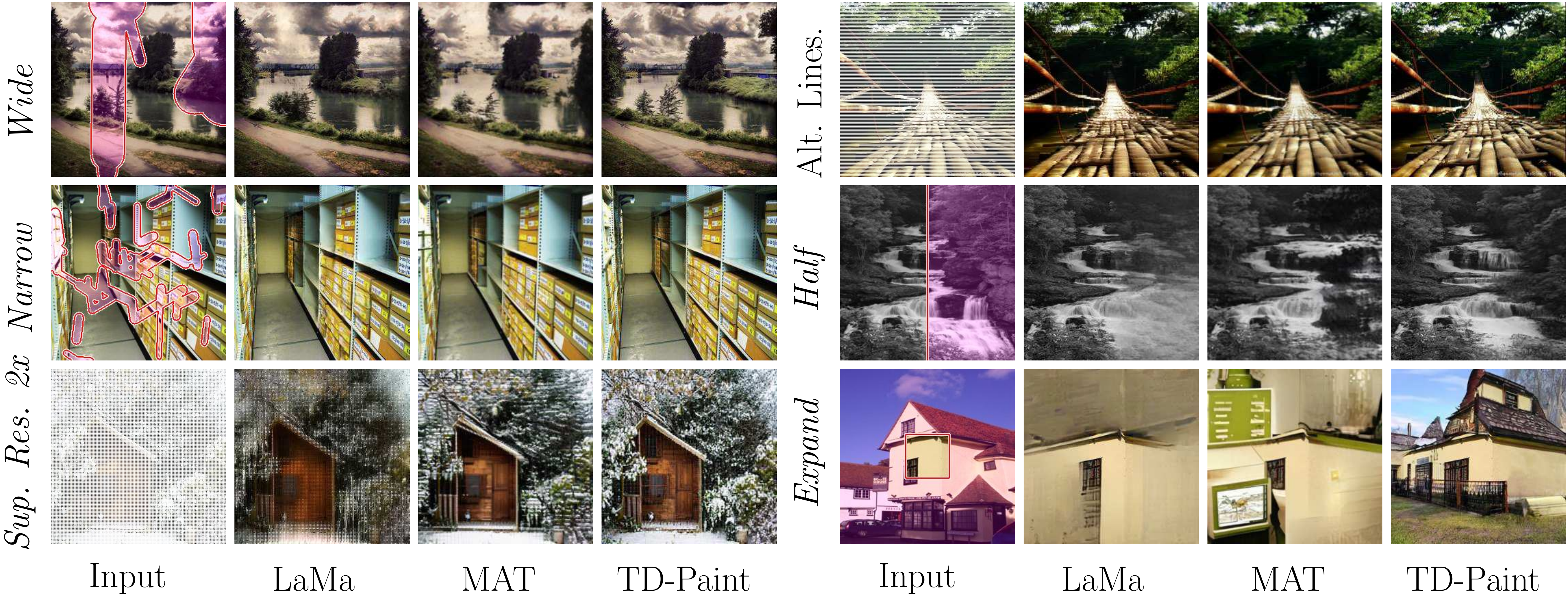}
    \caption{
    \textbf{Qualitative results:} \model against state-of-the-art inpainting CNN- and transformer-based models on \places. 
    Additional examples can be found in \cref{appendix:moreimages}.
    }
    \label{fig:exp:places:others}
\end{figure}

\noindent\textbf{\twotime and \altline masks}
\cref{table:exp:celebahq_imagenet:gan_transformer} shows that \model outperforms the baselines by a wide margin. Particularly \modelmat struggles with this task and often produces images with significant artifacts and blurring (see \cref{fig:exp:celebahq:others,fig:exp:imagenet:others,fig:exp:places:others}). In contrast, \model consistently achieves lower \lpips scores and higher \ssim values, reflecting superior image quality and performance.

\noindent\textbf{\half and \expand masks}
On \celebahq, \model achieves the best results across all datasets, as indicated by the \kid metrics (see \cref{table:exp:kid_table:others}).
As shown in \cref{fig:exp:celebahq:others,fig:exp:imagenet:others,fig:exp:places:others}, \modellama generates blurry artifacts in both the \half and \expand settings, whereas our proposed model consistently produces high-quality images.
This behavior of \modellama may be due to overfitting to the training mask distribution, as suggested by previous studies \citep{lugmayr2022repaint}.

\begin{table*}[!b]
    \centering
    \caption{
    \textbf{Quantitative results:} evaluation of CNN- and transformer-based models for inpainting on the \celebahq, \imagenet and \places datasets.}
    \includegraphics[width=1\contentwidth]{tables/celebahq_and_imagenet_GAN_Transformer.tex}
    \label{table:exp:celebahq_imagenet:gan_transformer}
\end{table*}

\begin{table*}[!b]
    \centering
    \caption{
    \textbf{Quantitative results:} \kid evaluation of CNN- and transformer-based models for inpainting on the \celebahq, \imagenet and \places datasets.}
    \includegraphics[width=1\contentwidth]{tables/kid_table_others.tex}
    \label{table:exp:kid_table:others}
\end{table*}

\section{Additional Qualitative Results}
\label{appendix:moreimages}
We provide additional qualitative inpainting results compared to the state-of-the-art models described in \cref{partie:experiments,vsother}.

\begin{figure}[th]
\centering
\includegraphics[width=0.98\contentwidth]{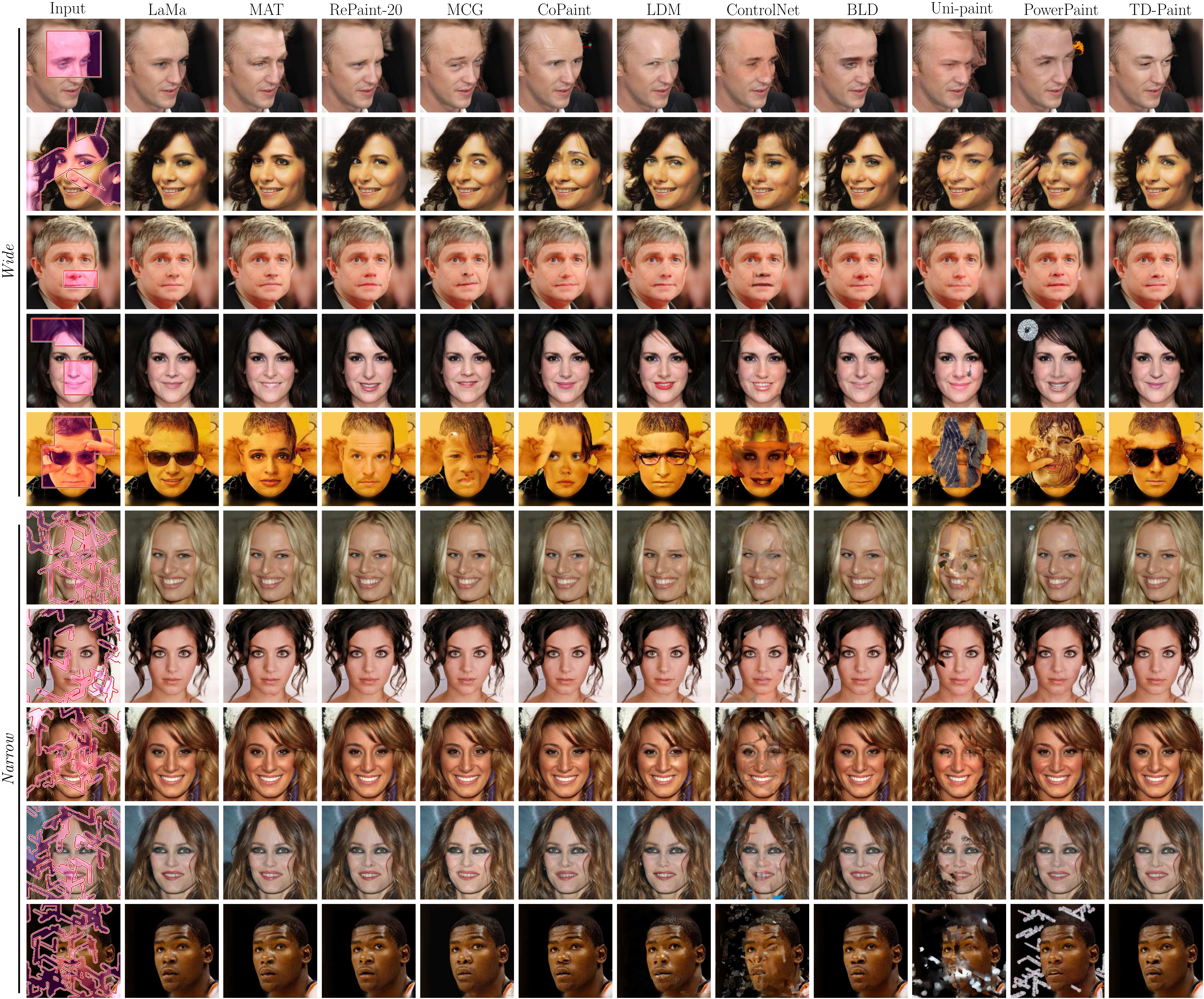}
\caption{\celebahq qualitative results}
\label{fig:appendix:celeba1}
\end{figure}

\begin{figure}[th]
\centering
\includegraphics[width=0.98\contentwidth]{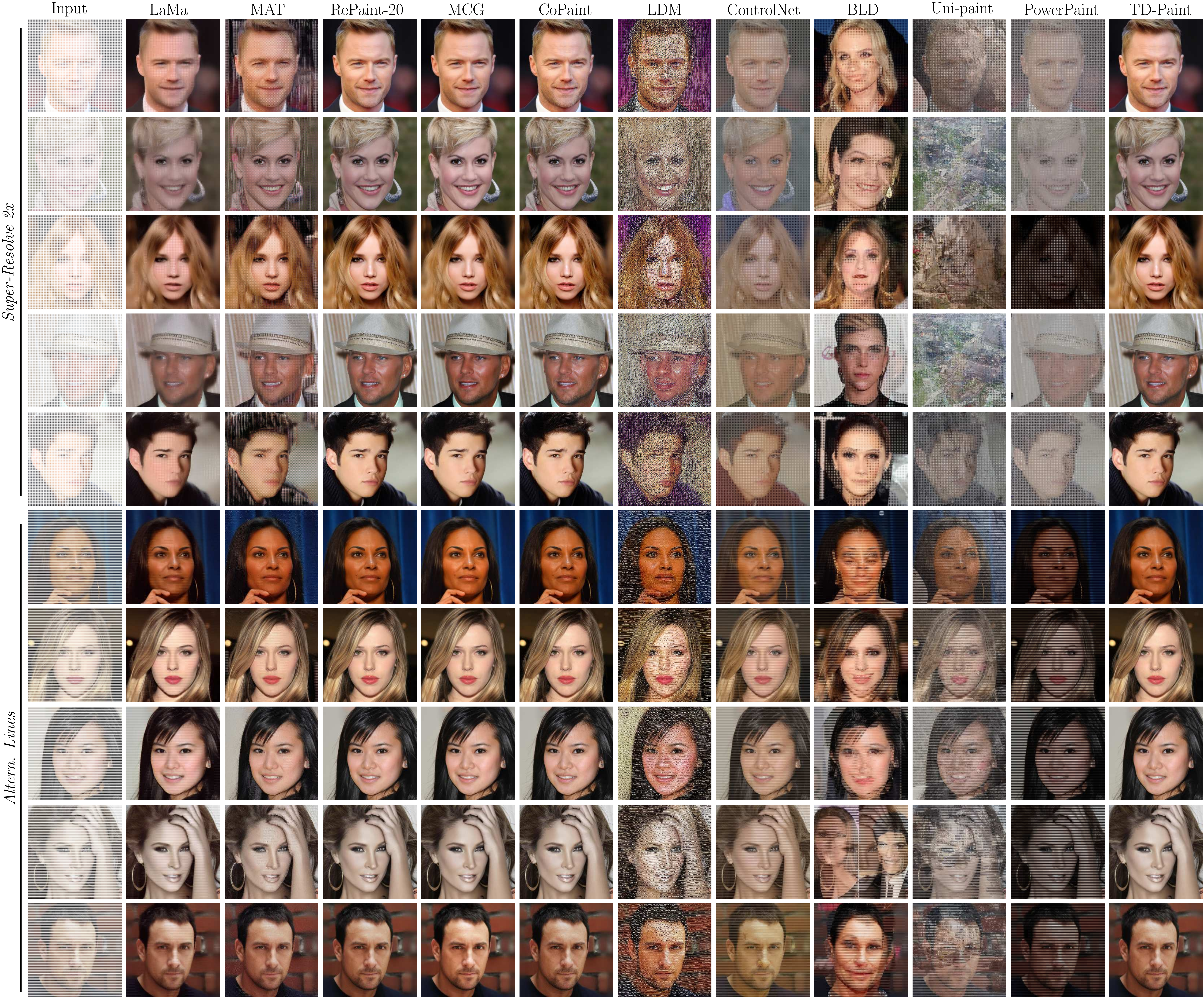}
\caption{\celebahq qualitative results}
\label{fig:appendix:celeba2}
\end{figure}

\begin{figure}[th]
\centering
\includegraphics[width=0.98\contentwidth]{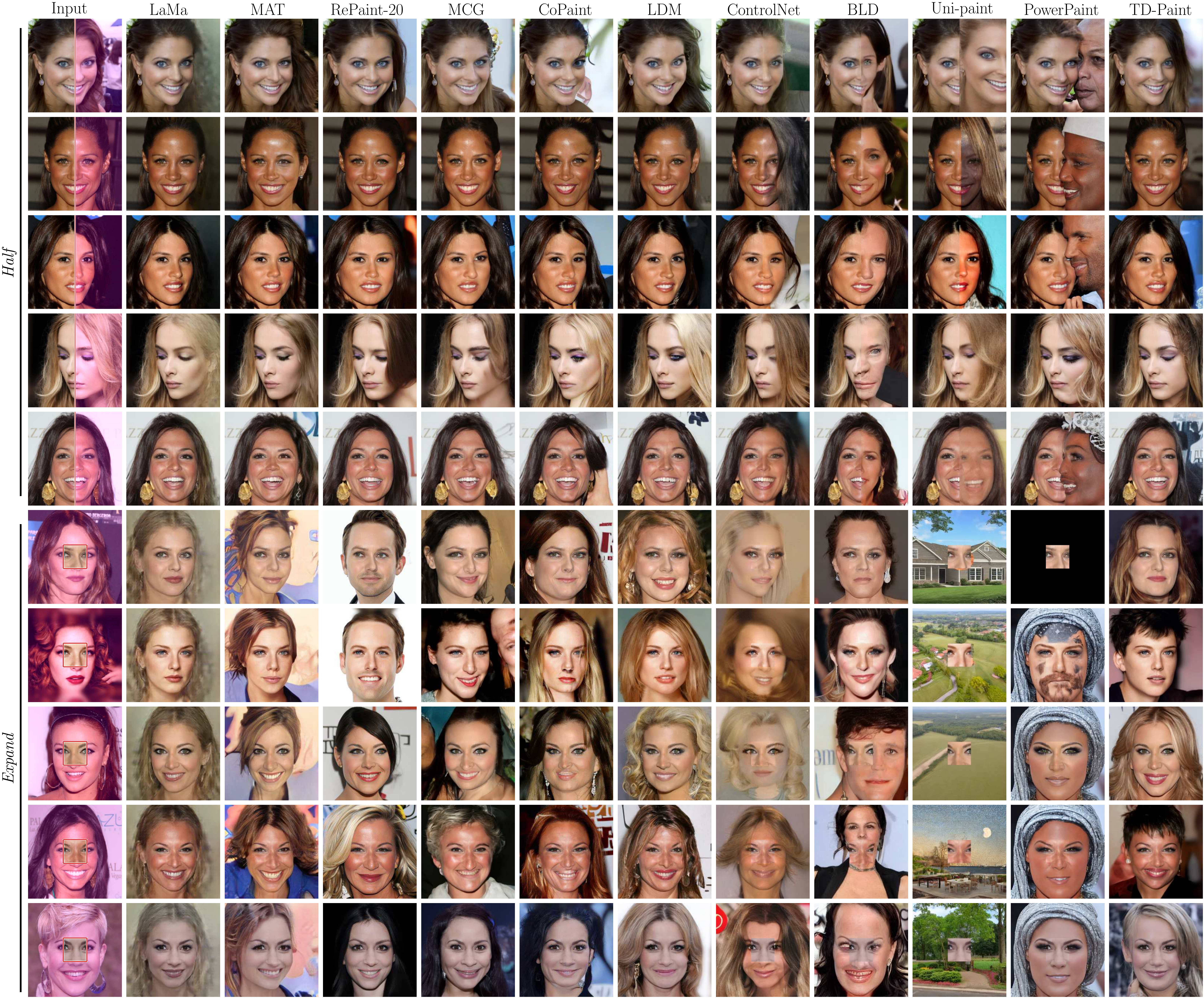}
\caption{\celebahq qualitative results}
\label{fig:appendix:celeba3}
\end{figure}

For \celebahq on 
\wide and \narrow masks in \cref{fig:appendix:celeba1},
\twotime and \altline masks in \cref{fig:appendix:celeba2},
\half and \expand in \cref{fig:appendix:celeba3}.

\begin{figure}[th]
\centering
\includegraphics[width=0.98\contentwidth]{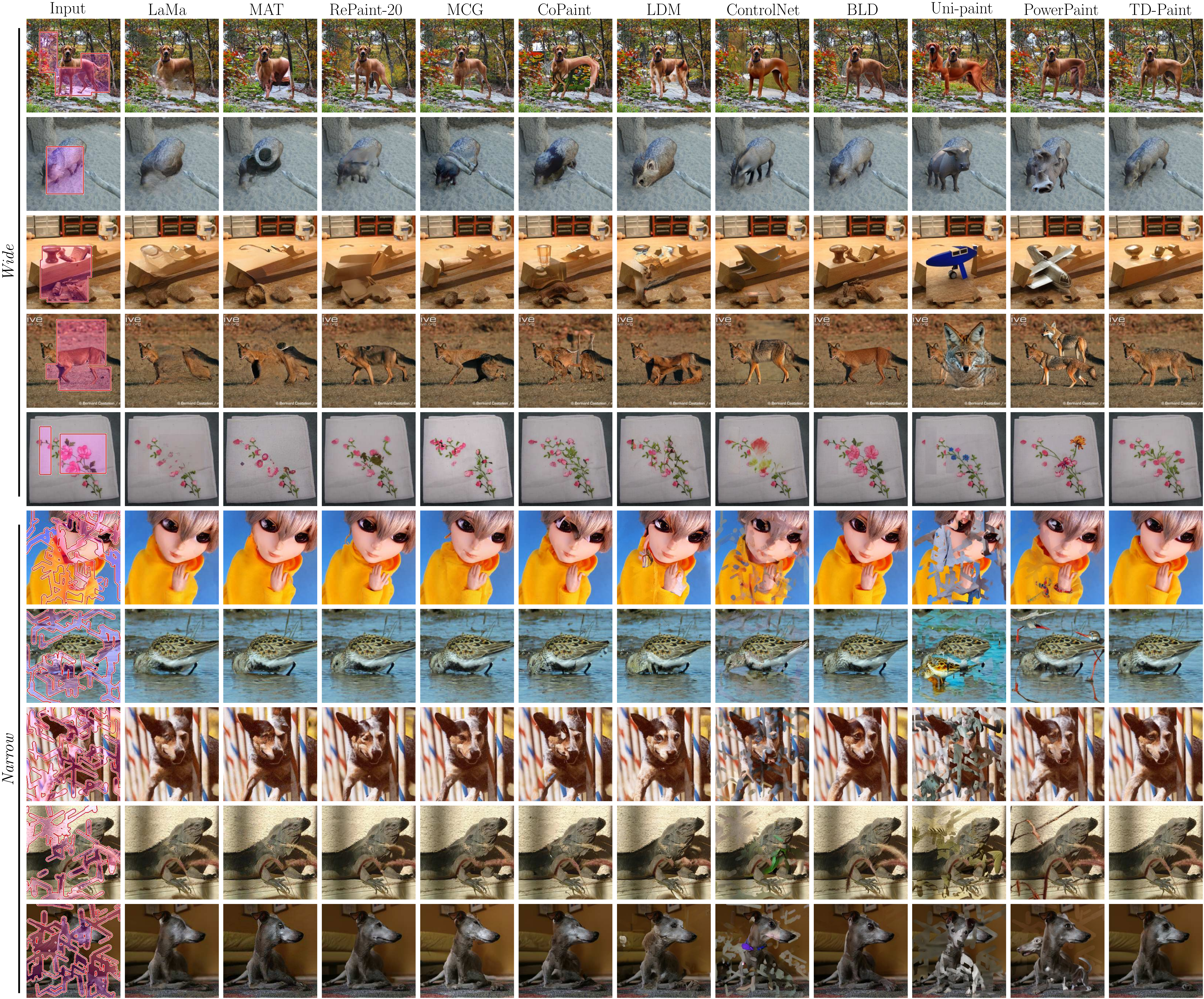}
\caption{\imagenet qualitative results}
\label{fig:appendix:imagenet1}
\end{figure}

\begin{figure}[th]
\centering
\includegraphics[width=0.98\contentwidth]{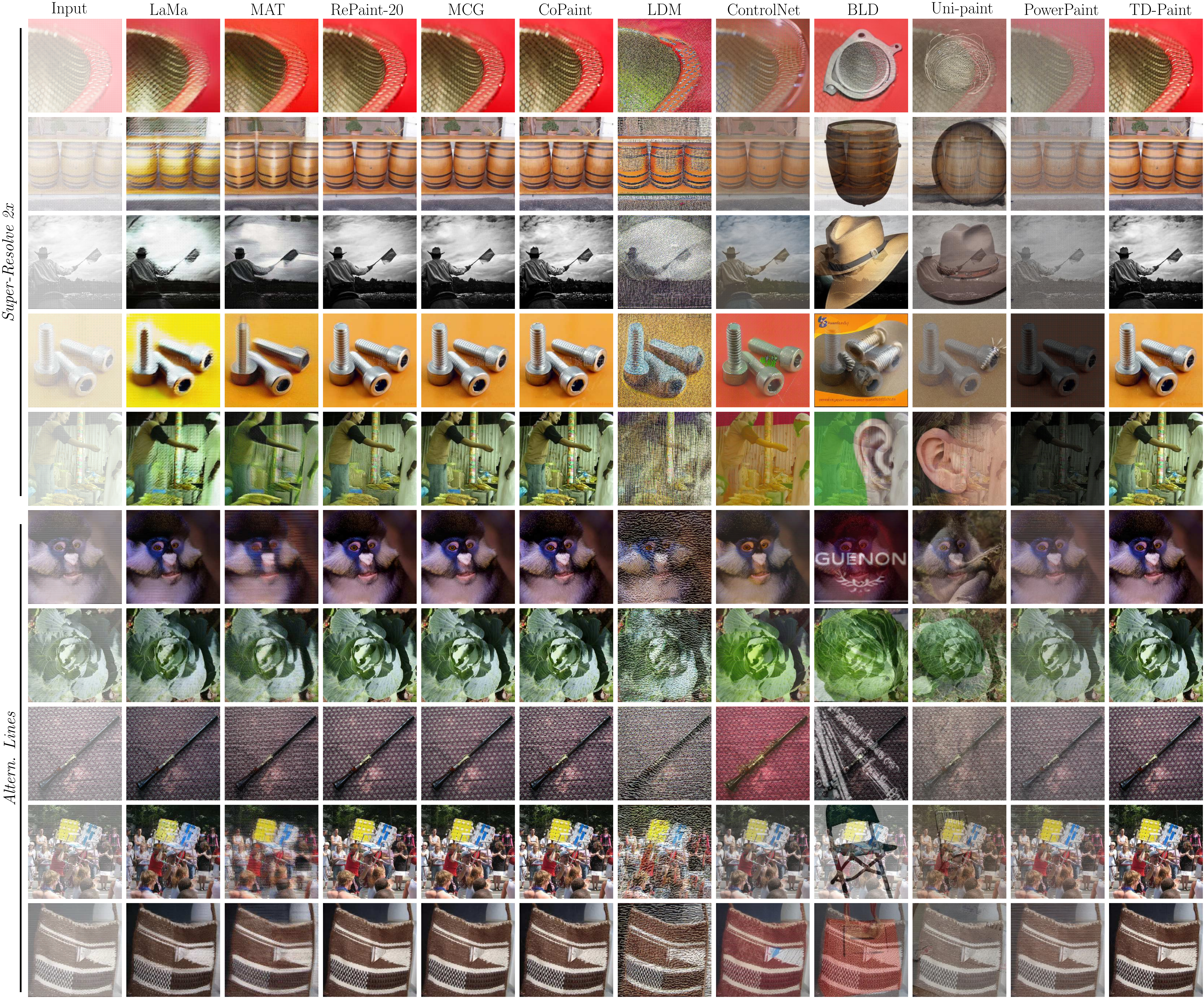}
\caption{\imagenet qualitative results}
\label{fig:appendix:imagenet2}
\end{figure}

\begin{figure}[th]
\centering
\includegraphics[width=0.98\contentwidth]{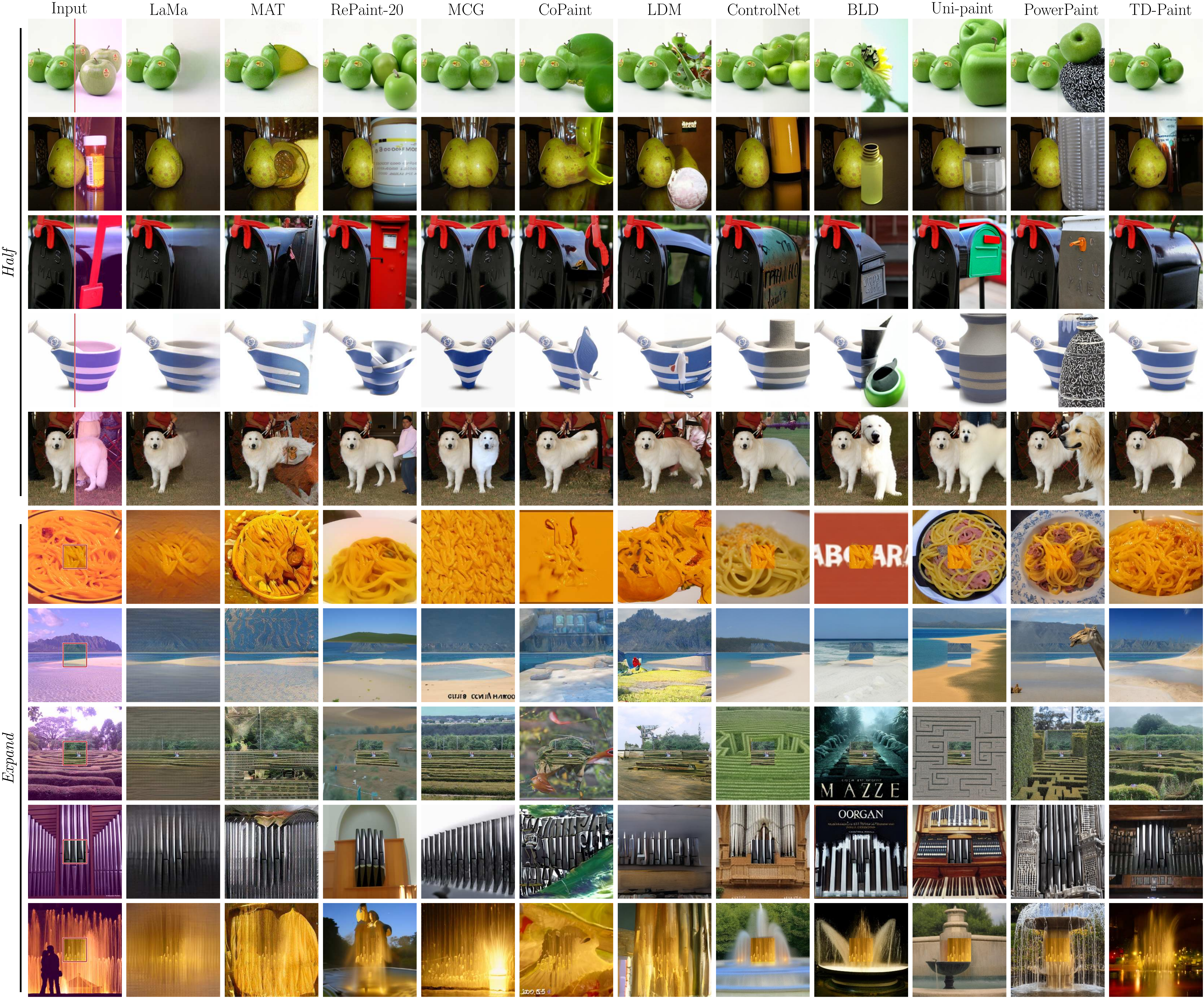}
\caption{\imagenet qualitative results}
\label{fig:appendix:imagenet3}
\end{figure}

For \imagenet on 
\wide and \narrow masks in \cref{fig:appendix:imagenet1},
\twotime and \altline masks in \cref{fig:appendix:imagenet2},
\half and \expand in \cref{fig:appendix:imagenet3}.

\begin{figure}[th]
\centering
\includegraphics[width=0.98\contentwidth]{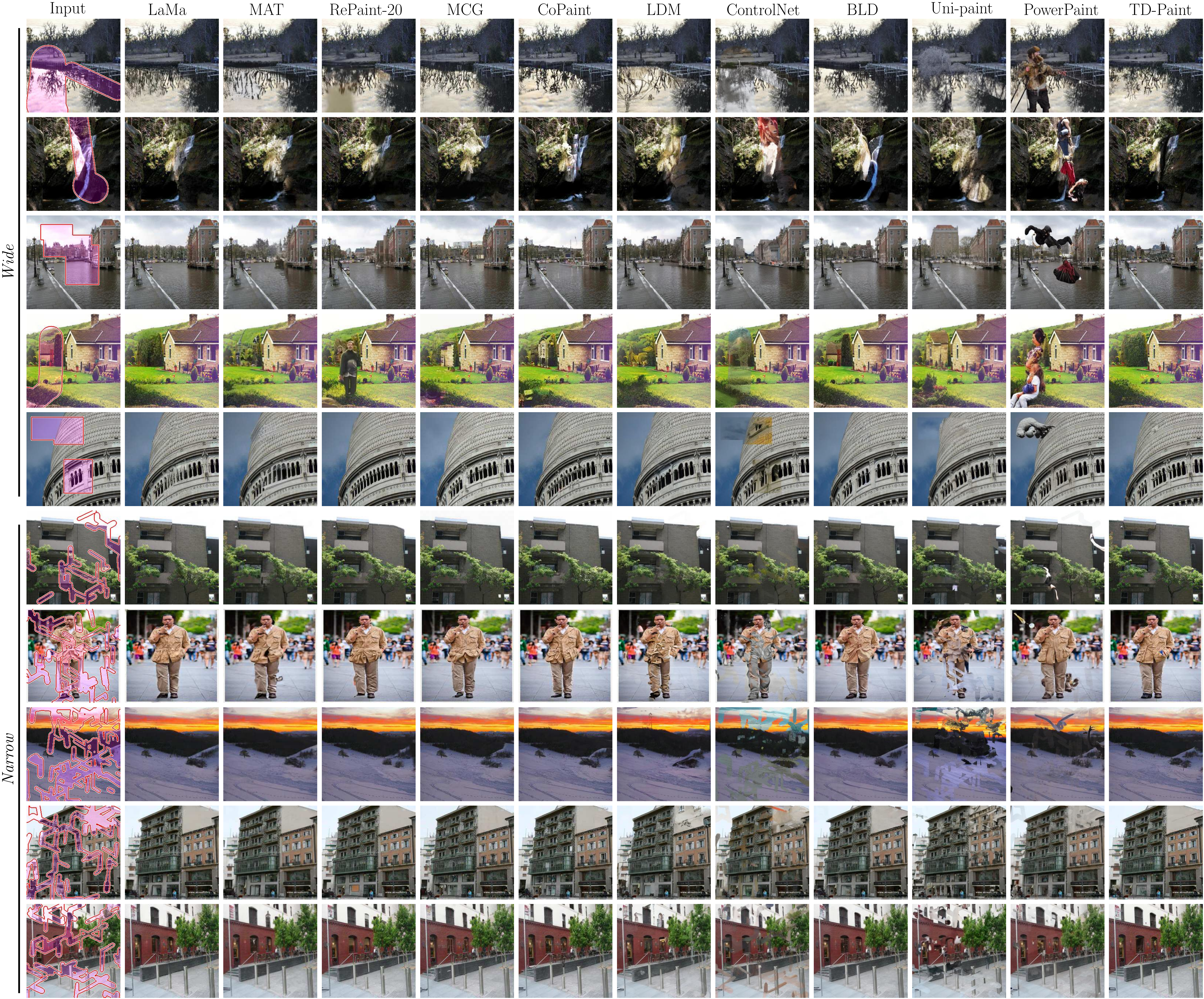}
\caption{\places qualitative results}
\label{fig:appendix:places1}
\end{figure}

\begin{figure}[th]
\centering
\includegraphics[width=0.98\contentwidth]{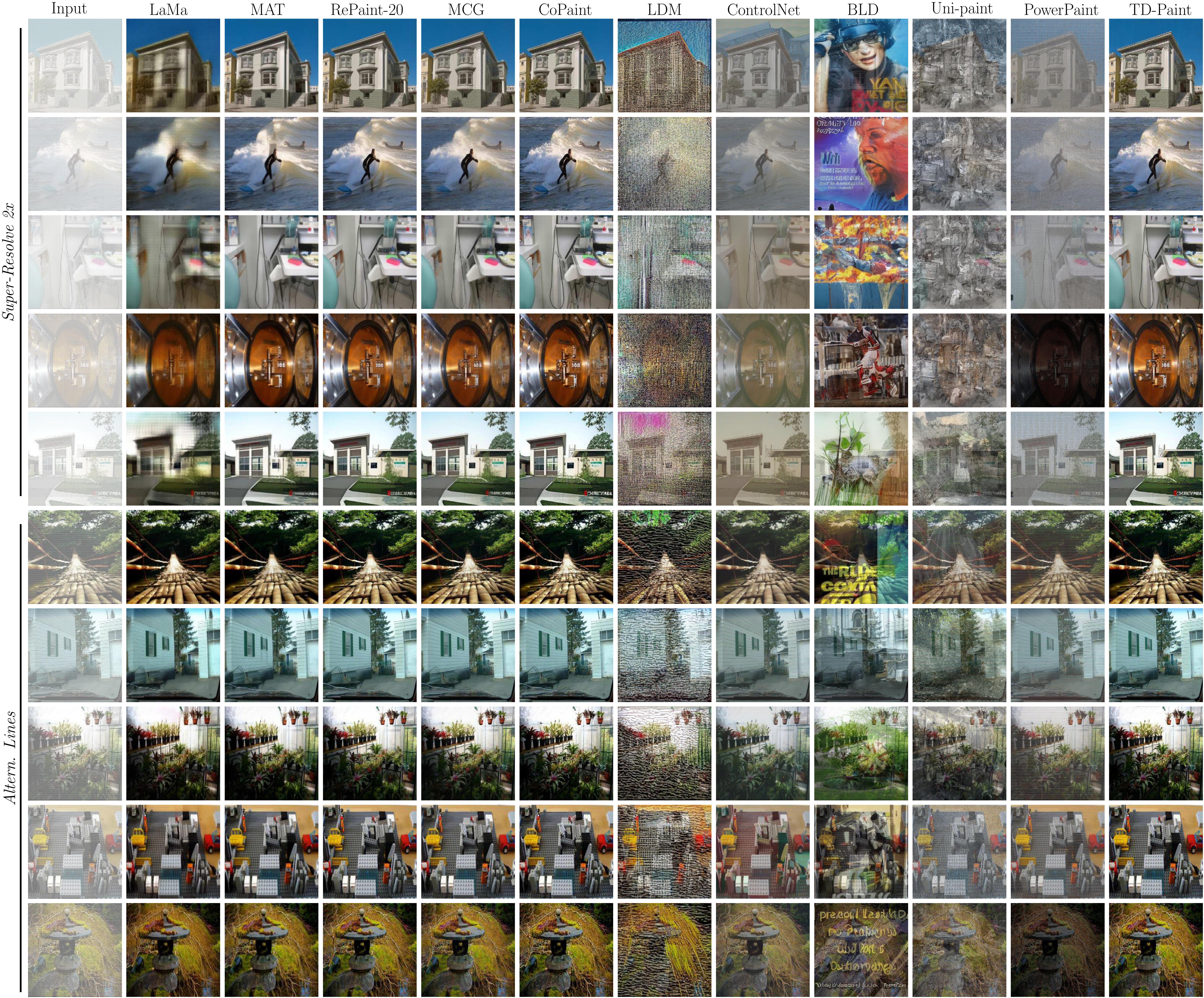}
\caption{\places qualitative results}
\label{fig:appendix:places2}
\end{figure}

\begin{figure}[th]
\centering
\includegraphics[width=0.98\contentwidth]{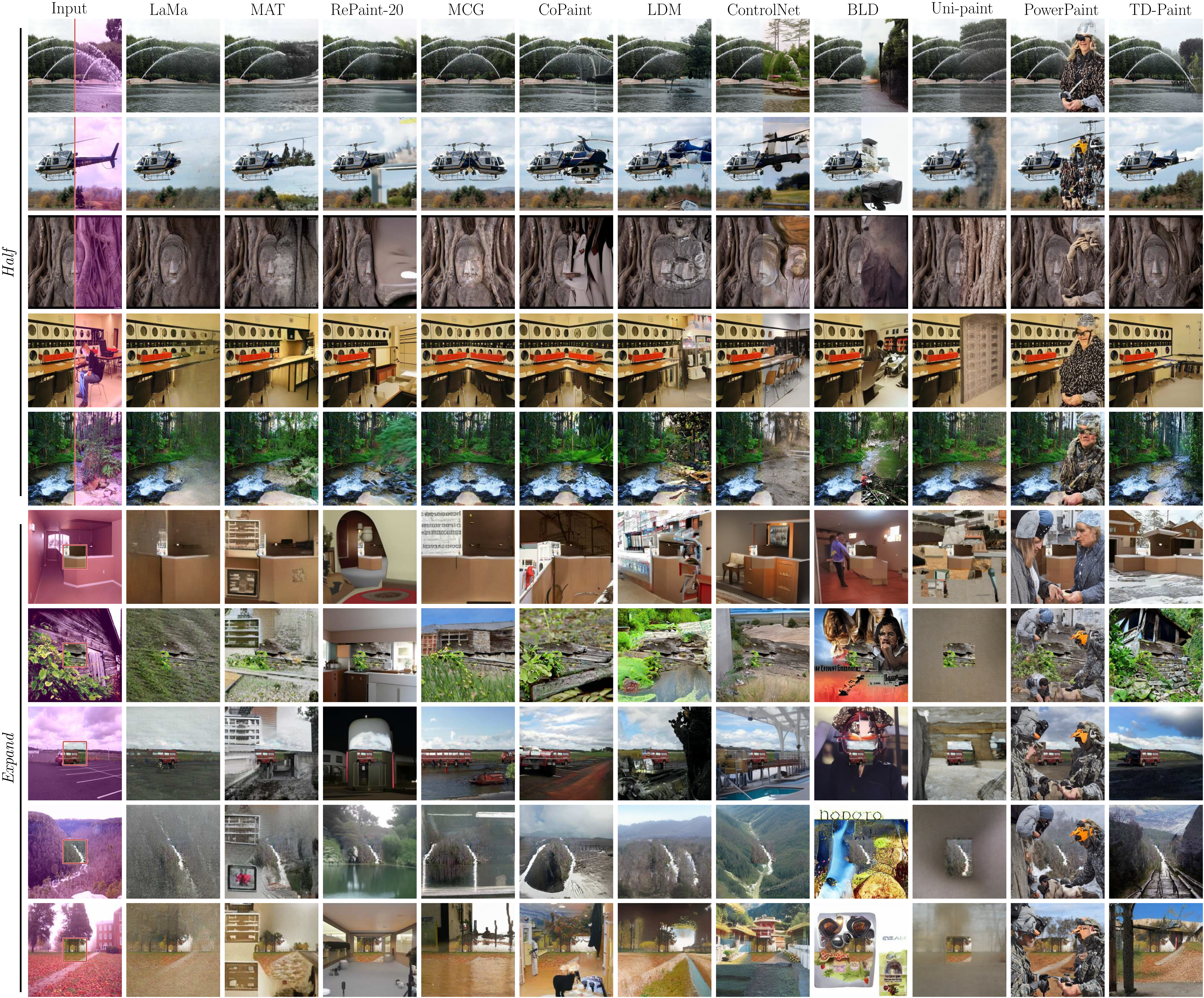}
\caption{\places qualitative results}
\label{fig:appendix:places3}
\end{figure}

For \places on 
\wide and \narrow masks in \cref{fig:appendix:places1},
\twotime and \altline masks in \cref{fig:appendix:places2},
\half and \expand in \cref{fig:appendix:places3}.

\begin{figure}[th]
\centering
\includegraphics[width=0.9\contentwidth]{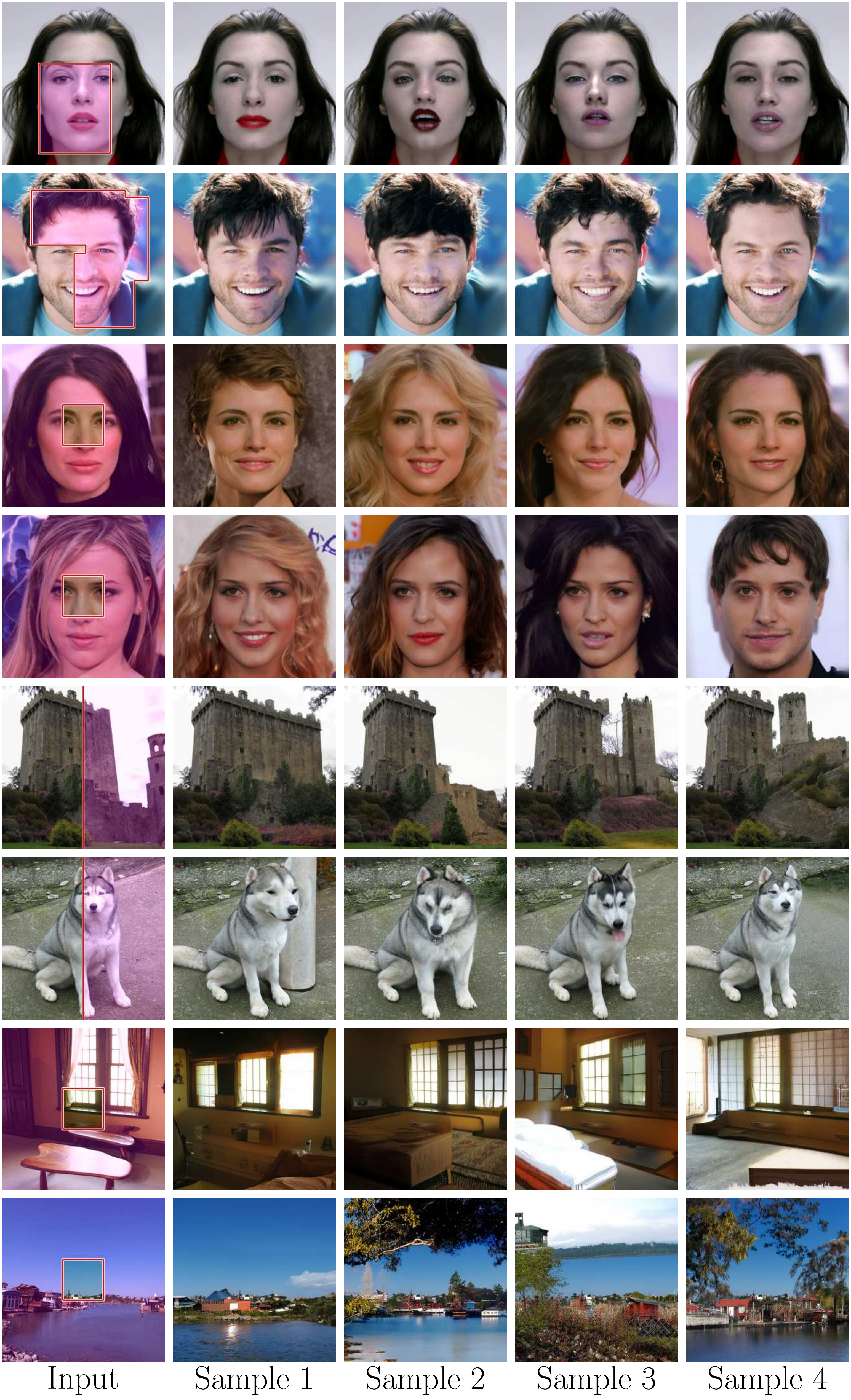}
\caption{
Example of divers generation using \model on \celebahq and \imagenet using the same input image and different initial noise.
}
\label{fig:exp:diversity:celebandimagenet}
\end{figure}

\begin{figure}[th]
    \centering
    \includegraphics[height=0.9\textheight]{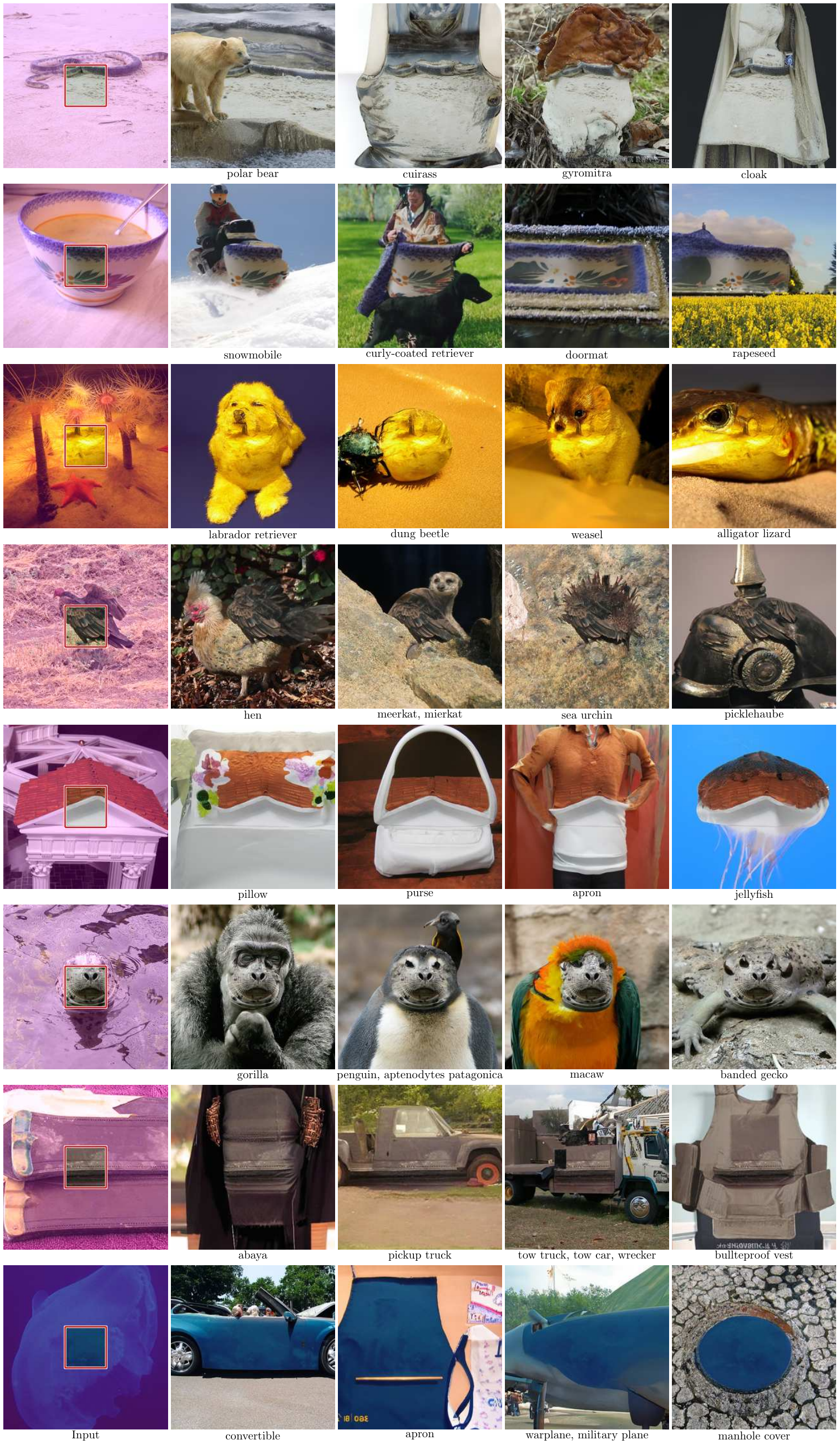}
    \caption{
    \imagenet \model diversity qualitative results using different class conditioning.
    For a line, \model is prompted with the same input image and mask but, with different classes.
    }
    \label{fig:appendix:imagenet:diversity_class}
\end{figure}

Additional diversity results in \celebahq and \imagenet can be found in \cref{fig:exp:diversity:celebandimagenet}, and for \imagenet with different conditional classes in \cref{fig:appendix:imagenet:diversity_class}. 
Through class conditioning, \model can generate diverse images based on different target classes. This mechanism allows \model guiding the inpainting process toward specific semantic categories, resulting in more controlled and contextually relevant image completions. This feature is especially useful when the inpainted region must match a particular class, increasing \model's flexibility and effectiveness across a range of inpainting tasks.

\begin{figure}[th]
    \centering
    \includegraphics[height=0.95\textheight]{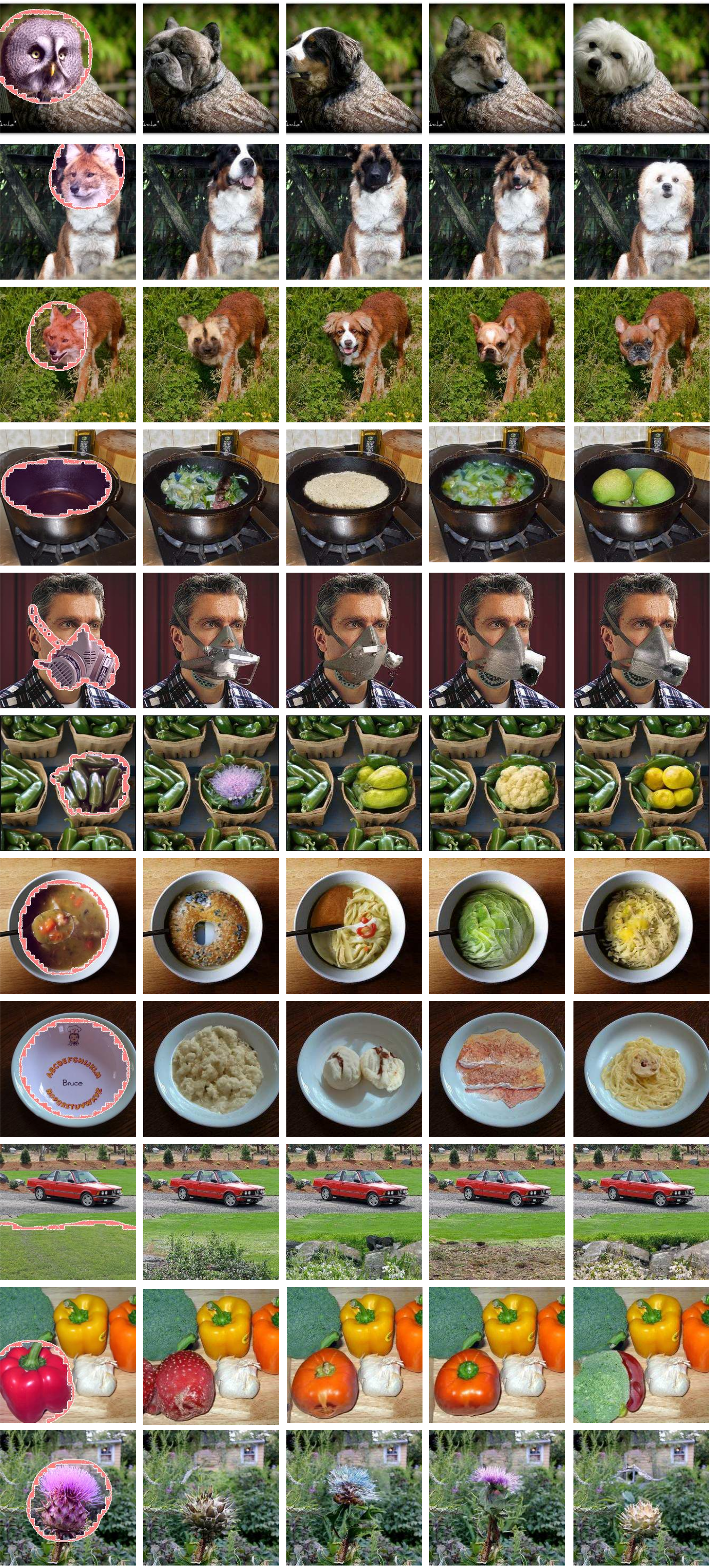}
    \caption{Demonstration of TD-paint application on the \imagenet dataset. The figure shows user-drawn masks highlighting specific regions or objects, followed by four generated image variations for each mask.}
    \label{fig:real_imagenet}
\end{figure}

\begin{figure}[th]
    \centering
    \includegraphics[height=0.95\textheight]{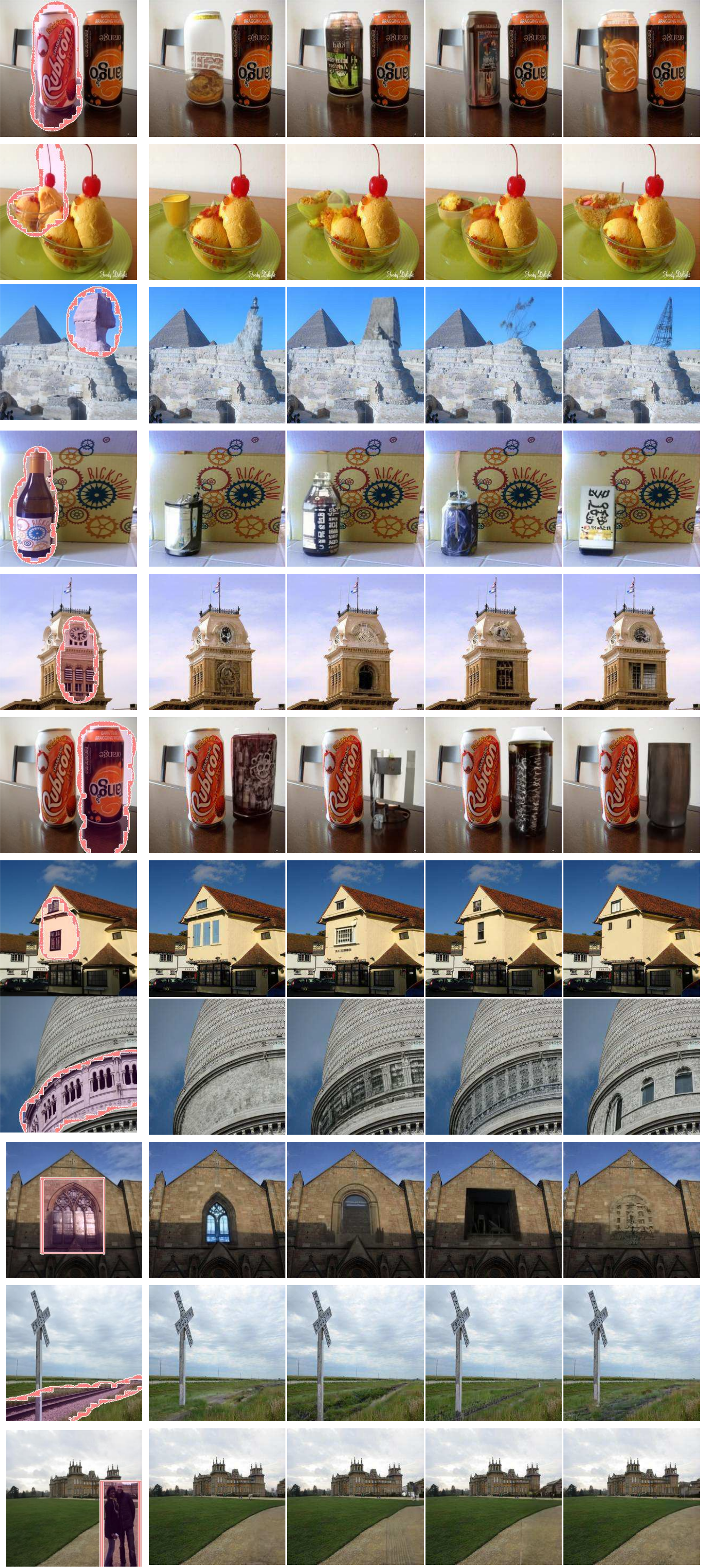}
    \caption{Demonstration of TD-paint application on the \places dataset. The figure shows user-drawn masks highlighting specific regions or objects, followed by four generated image variations for each mask.}
    \label{fig:real_places}
\end{figure}

Additional examples of inpainting using object-focused and region-specific masks are presented in \cref{fig:real_imagenet,fig:real_places}. These examples feature user-drawn masks that naturally follow object boundaries and regional structures, demonstrating \model's effectiveness in practical image manipulation scenarios. Such realistic mask shapes better reflect how users interact with inpainting tools in real-world applications, where selections typically correspond to meaningful objects or regions rather than arbitrary geometric patterns.

\newcommand{\appimgsize}{0.55}

\end{document}